\documentclass{article}

     \PassOptionsToPackage{numbers, compress}{natbib}

\usepackage[preprint]{neurips_2019}

\usepackage{placeins}
\usepackage{graphicx, color}
\usepackage{epstopdf}
\usepackage{latexsym}
\usepackage{amssymb}
\usepackage{epsfig, psfrag}
\usepackage{amsthm} 
\usepackage{graphics}
\usepackage{amsmath}
\usepackage{url}
\usepackage{xspace}
\usepackage[bookmarks]{hyperref}
\usepackage{fancyhdr}
\usepackage{caption}
\usepackage{subcaption}

\usepackage[utf8]{inputenc} 
\usepackage[T1]{fontenc}    
\usepackage{booktabs}       
\usepackage{amsfonts}       
\usepackage{nicefrac}       
\usepackage{microtype}      
\usepackage{courier}
\usepackage{floatrow}
\usepackage{mathtools}

\DeclarePairedDelimiter\floor{\lfloor}{\rfloor}

\newtheorem{theorem}{Theorem}
\newtheorem{lemma}{Lemma}

\newtheorem{assumption}{Assumption}

\theoremstyle{definition}
\newtheorem{definition} {Definition}
\newtheorem{remarks}{Remark}
\newtheorem{example}{Example}

\newfont{\boldlarge}{msbm10 scaled 1100}

\newcommand{\dist}[3]{\ell_{#1}\left(#2\mid #3\right)}
\newcommand{\samp}[2]{Y_{#1}^{(#2)}}

\newcommand{\est}[2]{b_{#1}^{(#2)}}

\newcommand{\ignore}[1]{}

\def\expe{\mathbb{E}}   
\def\argmin{\mathop{\rm argmin}}
\def\argmax{\mathop{\rm argmax}}

\def\P{\mathsf{P}}

\newcommand{\kl}[2]{D_{KL}\left(  #1   \left\| #2 \right. \right)}

\usepackage{mathtools}
\mathtoolsset{showonlyrefs=true}

\def\mc{\mathcal}
\def\mbf{\mathbf}
\def\mbb{\mathbb}
\def\mbs{\boldsymbol}

\newcommand{\yl}[1]{{\color{magenta}{YL: #1}}}

\title{Decentralized Bayesian Learning over Graphs}

\author{%
   Anusha Lalitha \qquad Xinghan Wang$^{\ast}$ \qquad Osman Kilinc~\thanks{denotes equal contribution} \qquad Yongxi Lu \\ \textbf{Tara Javidi} \qquad \textbf{Farinaz Koushanfar}\\
   University of California, San Diego \\
   \texttt{\{alalitha, x2wang, okilinc, yol070, tjavidi, fkoushanfar\}@eng.ucsd.edu}
}
\begin{document}
\allowdisplaybreaks
\maketitle

\begin{abstract}
We propose a decentralized learning algorithm over a general social network. The algorithm leaves the training data distributed on the mobile devices while utilizing a peer to peer model aggregation method. The proposed algorithm allows agents with local data to learn a shared model explaining the global training data in a decentralized fashion. The proposed algorithm can be viewed as a Bayesian and peer-to-peer variant of federated learning in which each agent keeps a \emph{"posterior probability distribution"} over a global model parameters. The agent update its "posterior" based on 1) the local training data and 2) the asynchronous communication and model aggregation with their 1-hop neighbors. This Bayesian formulation allows for a systematic treatment of model aggregation over any arbitrary connected graph. Furthermore, it provides strong analytic guarantees on converge in the realizable case as well as a closed form characterization of the rate of convergence. We also show that our methodology can be combined with efficient Bayesian inference techniques to train Bayesian neural networks in a decentralized manner. By empirical studies we show that our theoretical analysis can guide the design of network/social interactions and data partitioning to achieve convergence.

\end{abstract}

\section{Introduction}


Personal edge devices can often use their locally observed data to learn machine learning models that improve the user experience on the device as well as on other devices.  However, the use of local data for learning globally rich machine learning models has to address two important challenges. Firstly, this type of localized data, in isolation from the data collected by other devices, is unlikely to be statistically sufficient to learn a global model. Secondly, there might be severe restrictions on sharing raw forms of personal/local data due to privacy and communication cost concerns. In light of these challenges and restrictions, an alternative approach has emerged which leaves the training data distributed on the edge devices while enabling the decentralized learning of a shared model. This alternative, known as \emph{Federated Learning}, is based on edge devices' periodic communication with a central (cloud-based) server responsible for iterative model aggregation. While addressing the privacy constraints on raw data sharing, and significantly reducing the communication overload as compared to synchronized stochastic gradient descent (SGD), this approach falls short in fully decentralizing the training procedure.  Many practical peer to peer networks are dynamic and a regular access to a fixed central server, which coordinates the learning across devices, is not always possible. Existing methods based on federated learning cannot handle such general networks where central server is absent. To summarize, some of the major challenges encountered in a fully decentralized learning paradigm are: (i)\textit{Statistical Insufficiency:} The local and individually observed data distributions are likely to be less rich than the global training set. For example, a subset of features associated with the global model may be missing locally. (ii)\textit{Restriction on Data Exchange:} Due to privacy concerns, agents do not share their raw training data with the neighbors. Furthermore, model parameter sharing has been shown to reduce the communication requirements significantly. (iii) \textit{Lack of Synchronization:} There may not be a single agent with whom every agent communicates which can synchronize the learning periodically. (iv) \textit{Localized Information Exchange:} Agents are likely to limit their interactions and information exchange to a small 
group of their peers which can be viewed as the 1-hop neighbors on the social network graph. Furthermore, information obtained from different peers might be viewed differently, requiring a heterogeneous model aggregation strategy.

\textbf{Contributions:} We consider a fully decentralized learning paradigm where agents iteratively update their models using local data and aggregate information from their neighbors to their local models.  In particular, we consider a learning rule where agents take a Bayesian-like approach via the introduction of a posterior distribution over a parameter space characterizing the unknown global model. Our theoretical and conceptual contributions are as follows: (i) Our decentralized learning rule generalizes a learning rule considered in the social literature~\cite{7172262, 7349151, 8359193} by restricting the posterior distribution to a predetermined family of distributions for computational tractability. (ii) We provide theoretical guarantee that each agent will eventually learn the true parameters associated with global model under mild assumptions. (iii) We provide analytical characterization of the rate of convergence of the posterior probability at each agent in the network as a function of network structure and local learning capacity. (iv) Unlike prior work, we allow a fully general network structure as long as it is strongly connected. As a consequence, our work provides first known theoretical guarantees on convergence for a Bayesian variant of federated learning.

In addition to our theoretical results we show that our methodology can be combined with efficient Bayesian inference techniques to train Bayesian neural networks in a decentralized manner. By empirical studies we show that our theoretical analysis can guide the design of network/social interaction and data partition to achieve convergence. We also show the scalability of our method by training over 100 neural networks on asynchronous time-varying networks. Our Bayesian approach has the added advantage of obtaining confidence value over agents' predictions and can directly benefit from Bayesian learning literature which shows that these models offer robustness to over-fitting, regularization of the weights, uncertainty/confidence estimation, and can easily learn from small datasets~\cite{gal2016uncertainty, local_repara_trick_kingma}. In this regard, our work bridges the gap between decentralized training methodologies and Bayesian neural networks.

\textbf{Related Work:} Our fully decentralized training methodology extends federated learning~\cite{DBLP:journals/corr/KonecnyMRR16, konevcny2016federated, mcmahan2017communication} to general graphs in a Bayesian setting and does away with the need of having a centralized controller. In particular, our learning rule also generalizes various Bayesian inference techniques such as~\cite{vcl_iclr, weight_uncert_NN, streaming_var_bayes, local_repara_trick_kingma} and variational continual learning techniques such as~\cite{vcl_iclr, streaming_var_bayes}. Lastly, our work can be viewed as a Bayesian variant of communication-efficient methods based on SGD~\cite{JMLR:v18:16-512, DBLP:journals/corr/ChaudhariBZST17, DBLP:journals/corr/abs-1808-07217} which also allow the agents to make several local computations and then periodically average the local models. This is unlike decentralized optimization and SGD based methods~\cite{Duchi2012DualAF, 6425904, pmlr-v80-tang18a, doi:10.1137/16M1080173, NIPS2017_7117, NIPS2017_7172, DBLP:journals/corr/JinYIK16} where local (stochastic) gradients are computed for each instance of data and communication happens at a rate comparable to number of local updates. For a detailed overview on the communication-efficient SGD methods contrasted with decentralized optimization methods refer to~\cite{DBLP:journals/corr/abs-1808-07576}.



\noindent \textbf{Notation}:  We use boldface for vectors $\mathbf{v}$  and denote its $i$-th element by $v_{i}$. Let $[n]= \{1, 2, \ldots, n\}$. Let $\mc{P}(A)$ and $| A|$ denote the set of all probability distributions and the number of elements resp. on a set $A$. Let $G(x, \theta, \sigma^2)$ denote the pdf of a Gaussian random variable with mean $\theta$ and variance $\alpha^2$. Let $D_{\text{KL}}(P_{Z}||P_Z')$ be the Kullback--Leibler (KL) divergence between two probability distributions $P_Z, P_Z' \in \mc{P}(\mc{Z})$. 


\section{Problem Formulation}

\paragraph{The Model:} Let $\mc{X}$ denote the global input space and let $\mc{Y}$ denote the set of all possible labels. The global dataset has input-label pairs belonging to $(\mc{X}, \mc{Y})$ which are distributed as $\mc{D} = \P_X \times \P_{Y|X}$. Consider a group of $N$ individual agents, where each agent $i$ has access to input-label pairs taken from a subset $(\mc{X}_i, \mc{Y})$ such that $\cup_{i=1}^N \mc{X}_i \subset \mc{X}$. The samples $\left\{X_i^{(1)}, X_i^{(2)}, \ldots\right\}$ are independent and identically distributed (i.i.d), and are generated according to the distribution $\P_i \in \mc{P}(\mc{X})$. Furthermore, we assume that each agent has a set of \textit{candidate local likelihood functions} over the label space which are  parametrized by $\theta \in \Theta$ and given by $\{\ell_i(y \mid \theta, x):  y \in \mc{Y}, \, \theta \in \Theta, \, x \in \mc{X} \}$. Each agent $i$ is aiming to learn a distribution over $\Theta$ which achieves the following
\begin{align}
\label{eq:objective_func}
    \inf_{\pi \in \mc{P}(\Theta)}\expe_{\P_{X}}\left[D_{\text{KL}}\left(\P_{Y|X}(\cdot|X)\left|\left|\int_{\Theta}\ell_i(\cdot| \theta, X)\pi(\theta)d\theta\right)\right.\right.\right].
\end{align}


Note that for any input $x \sim \P_X$, the distribution $\int_{\Theta}\ell_i(\cdot| \theta, x)\pi(\theta)d\theta$ denotes predictive distribution over the label space $\mc{Y}$. Minimizing the objective in equation~\eqref{eq:objective_func} ensures that each agent makes statistically similar predictions as the true labelling function over the global dataset.    

\ignore{The distribution that minimizes equation~\eqref{eq:objective_func} at each agent $i$ is a function of the (prior on the) candidate local likelihood functions, the local dataset seen by agent $i$ and the information obtained from the neighbors. }


\begin{definition}
A social learning model is said to be \textit{realizable} 
if there exists a $\theta^{\ast} \in \Theta$ such that $ \ell_i(\cdot \mid \theta^{\ast}, x) = \P_{Y|X}(\cdot\mid x)$ for $i\in [N]$. 
\end{definition}

We note that, in the realizable case, the minimizer of equation~\eqref{eq:objective_func} is the trivial distribution which takes value one at $\theta^{\ast}$ and zero elsewhere. In other words, in the realizable case, each agent's goal is to learn the \emph{true model parameter} $\theta^{\ast}$. 

\begin{definition}
If $\P_i = \P_X$ for all agents $i \in [N]$, then all agents have identically distributed observations across the network. We refer to this as the \textit{IID data distribution} setting. In contrast, we call the local data to have 
\textit{non-IID data distribution} when there exists $i \in [N]$ for which $\P_i \neq \P_X$.  
\end{definition}


\begin{example}[Decentralized Linear Regression with non-IID Data Distribution]
\label{ex:dist_linear_regression}
Let $d \geq 2$ and $\Theta = \mbb{R}^{d}$. Consider a linear realizable model where there exists a $\mbs{\theta}^{\ast} = [\theta_0^{\ast}, \ldots, \theta^{\ast}_{d-1}]\in \Theta$, data input $\mbf{x} \in \mbb{R}^d$, the label $y \in \mbb{R}$ given as 
$
y = {\mbs{\theta}^{\ast}}^T\mbs{\phi}(\mbf{x}) + \eta,
$
where the basis function $\mbs{\phi}:\mbb{R}^{d}:\to \mbb{R}$ provides the feature vector $\mbs{\phi}(\mbf{x}) = [\phi_0(\mbf{x}), \ldots,  \phi_{d-1} (\mbf{x})]^T$ and $\eta$ denotes the additive Gaussian noise $\eta \sim N(0, \alpha^2)$. This implies true probabilistic model generating the labels as well as local likelihood function at any agent $i$, given an input $\mbf{x}$ is given by $\P_{Y|X}(y \mid \mbf{x}) = \ell_i(y \mid \mbs{\theta}, \mbf{x}) = G(y, {\mbs{\theta}^{\ast}}^T\mbs{\phi}(\mbf{x}), \alpha^2 )$. Now we consider a non-IID data distribution: Fix some $0<m < d$ and let $\mc{X}_1 = \left\{\mbf{x} \in \mbb{R}^{d+1} \mid \mbs{\phi}(\mbf{x}) = [\phi_0(\mbf{x}), \ldots,  \phi_{m-1} (\mbf{x}), 0, \ldots, 0]^T \right\}$ and $\mc{X}_2 = \left\{\mbf{x} \in \mbb{R}^{d+1}\mid \mbs{\phi}(\mbf{x}) = [0, \ldots, 0,  \phi_m (\mbf{x}), \ldots,  \phi_{d-1} (\mbf{x})]^T\right\}$. Suppose that agent 1 make observations in $\mc{X}_1$ or can access only $m$ features locally. Similarly, agent 2 observations lie in $\mc{X}_2$, i.e. the remaining $d-m$ features locally. It is clear that the local features at each agent is such that the true parameter $\mbs{\theta}^{\ast}$
cannot be locally learned and there is a need for communication and model aggregation. 


\end{example}


\begin{example}[Decentralized Image Classification using Deep Neural Networks] 
\label{ex:dist_dnn}
Consider the problem of learning a neural network which can approximate the input-label probabilistic model with distribution $\P_{Y|X}(\cdot\mid \mbf{x})$ over the label space for each input image $\mbf{x}\in \mc{X}$. 
In this setting, the local likelihood function at any agent $i$, given an image $\mbf{x} \in \mc{X}$ was observed, conditioned on the DNN weights $\mbs{\theta}$ is obtained as follows
$
    \ell_i(y \mid \mbs{\theta}, \mbf{x})
    = 
    \text{Softmax}(y,\mbf{f}_{\mbs{\theta}}(\mbf{x}))
    :=
    \frac{\exp(\mbf{f}^y_{\mbs{\theta}}(\mbf{x}))}{\sum_{y^{\prime} \in \mc{Y}}\exp(\mbf{f}^{y^{\prime}}_{\mbs{\theta}}(\mbf{x}))},
$
where $\mbf{f}^y_{\mbs{\theta}}(\cdot)$ denotes the value of the output layer of the neural network at label $y$.


\end{example}

\paragraph{The Communication Network:}
We model the communication network between agents via a directed graph with vertex set $[N]$. We define the neighborhood of agent $i$, denoted by $\mathcal{N}(i)$, as the set of all agents $j$ who have an edge going from $j$ to $i$. We assume $i \in \mc{N}(i)$. Furthermore, if agent $j \in \mc{N}(i)$, agent $i$ receives information from agent $j$.  The social interaction of the agents is characterized by a stochastic matrix $W$. The weight $W_{ij} \in [0, 1]$ is strictly positive if and only if $j \in \mc{N}(i)$ and  $\sum_{j =1}^N W_{ij} = 1$. The weight $W_{ij}$ denotes the confidence agent $i$ has on the information it receives from agent $j$.


\subsection{Decentralized Learning Rule}
\label{sec:algorithm}



We introduce a decentralized learning rule which generalizes a learning rule considered in the social learning literature~\cite{7172262, 7349151, 8359193}. However, we restrict local posterior distributions to a predetermined family of distributions. This allows us to implement the decentralized algorithm in a computationally tractable manner.
Let $\mc{Q} \subset \mc{P}(\Theta)$ a family of posterior distributions. Start with $\mbf{q}_{i}^{(0)} \in \mc{P}(\Theta)$ with $\mbf{q}_{i}^{(0)}(\theta) > 0$ for all $\theta \in \Theta$ and $i \in [N]$. At each time step
$n = 1, 2, \ldots$ the following events happen at every agent $i \in [N]$:

\begin{enumerate}
\item
Draw a batch of $M$ i.i.d samples $\left(\mbf{X}^{(n)}_i,\mbf{Y}^{(n)}_i\right) \sim \P_{Y|X}(\mbf{Y}^{(n)}_i|\mbf{X}^{(n)}_i)\P^M_i(\mbf{X}^{(n)}_i)$.

\item
\textbf{Local Bayesian Update of Posterior:} Perform a local Bayesian update on $\mbf{q}_{i}^{(n-1)}$ to form the public posterior vector $\mbf{b}_{i}^{(n)}$ using the following rule.
	 For each $\theta \in \Theta$, 
		\begin{align}
		{b}_{i}^{(n)} (\theta) &=\frac{ \ell_i\left(\mbf{Y}_{i}^{(n)} \mid \theta, \mbf{X}_{i}^{(n)}\right) q_{i}^{(n-1)}(\theta) }{ \int_{\Theta} \ell_i\left(\mbf{Y}_{i}^{(n)} \mid \phi, \mbf{X}_{i}^{(n)}\right) q_i^{(n-1)}(\phi) d\phi }.
		\label{eq:bayes}
		\end{align}
		
\item
\textbf{Projection onto Allowed Family of Posteriors:} Project onto an allowed family of posterior distributions $\mc{Q}$ by employing KL-divergence minimization,
\begin{align}
\vspace{-0.2cm}
\label{eq:proj_step}
\Pi_{\mc{Q}}(\mbf{b}_{i}^{(n)}) = \argmin_{\pi \in \mc{Q}} \kl{\pi}{\mbf{b}_{i}^{(n)}}.
\end{align}

\item
\textbf{Communication Step:} Agent $i$ sends $\Pi_{\mc{Q}}(\mbf{b}_{i}^{(n)})$ to agent $j$ if $i \in \mc{N}(j)$ and receives $\Pi_{\mc{Q}}(\mbf{b}_{j}^{(n)})$ from neighbors $j \in \mc{N}(i)$.
	
\item
\textbf{Consensus Step:} Update private posterior distribution by averaging the log posterior distributions received from neighbors, i.e., for each  $\theta \in \Theta$,
		\begin{align}
		\label{eq:consensus}
		\vspace{-0.2cm}
		q_{i}^{(n)}(\theta) = \frac{ \exp \left( \sum_{j = 1}^{N} W_{ij} \log \Pi_{\mc{Q}}(\mbf{b}_{j}^{(n)})(\theta) \right)
			}{
			\int_{\Theta} \exp \left( \sum_{j = 1}^{N} W_{ij} \log \Pi_{\mc{Q}}(\mbf{b}_{j}^{(n)})(\phi) \right)d\phi
			}.
		\end{align}
\end{enumerate}



\begin{remarks}[Variational Inference]
\label{remm:VI}
In above learning rule, local Bayesian update of the posterior step~\eqref{eq:bayes} can be combined with the projection onto allowed family of distributions~\eqref{eq:proj_step} as follows
\begin{align}
\vspace{-0.2cm}
\mbf{b}_{i}^{(n)} 
& = \argmin_{\pi \in \mc{Q}} \kl{\pi}{\frac{1}{Z^{(n)}_i} \ell_i\left(\mbf{Y}_{i}^{(n)} \mid \cdot, \mbf{X}^{(n)}_i\right) \est{i}{n-1}(\cdot)}
\\
& = \argmin_{\pi \in \mc{Q}} \kl{\pi}{\mbf{q}_{i}^{(n-1)}} + \expe_{\pi}\left[-\log \ell_i\left(\mbf{Y}_{i}^{(n)} \mid \cdot, \mbf{X}^{(n)}_i\right)  \right],\label{eq:variational_energy_min}
\end{align}
where $Z_i^{(n)}$ is the possibly intractable normalization constant. Minimization performed in Equation~\eqref{eq:variational_energy_min} is referred to as Variational Inference (VI) and the minimand is referred to as the variational free energy~\cite{murphy_ml_book, weight_uncert_NN, local_repara_trick_kingma, gal2016uncertainty}. 
\end{remarks}


\begin{remarks}[Gaussian Approximate Posterior]
Gaussian approximate posterior can be obtained in an computationally efficient manner via VI techniques~\cite{weight_uncert_NN, local_repara_trick_kingma}. More specifically, let $\mc{Q}$ denote the family of Gaussian posterior distributions with pdf given by $G(\mbs{\theta}, \mbs{\mu}, \mbs{\Sigma})$. Let $(\mbs{\mu}^{(n)}_i, \mbs{\Sigma}_i^{(n)})$ denote the mean and the covariance matrix of $\mbf{b}_i^{(n)}$ at agent at $i$ obtained using equation~\eqref{eq:variational_energy_min}. Then we can show that the posterior distribution $\mbf{q}_i^{(n)}$ obtained after the consensus step also belongs to $\mc{Q}$ for all $i \in [N]$. Furthermore, the mean and covariance matrix $(\mbs{\widetilde{\mu}}^{(n)}_i, \mbs{\widetilde{\Sigma}}_i^{(n)})$ of $\mbf{q}_i^{(n)}$ is given as follows
\begin{align}
\label{eq:gaussian_consensus_step}
    {\mbs{\widetilde{\Sigma}}_i^{(n)}}^{-1} 
    = \sum_{j=1}^{N}  W_{ij}{\mbs{\Sigma}_j^{(n)}}^{-1},
    \quad 
    \mbs{\widetilde{\mu}}^{(n)}_i
    = \mbs{\widetilde{\Sigma}}^{(n)}_i\sum_{j=1}^{N} W_{ij}{\mbs{\Sigma}^{(n)}_j}^{-1}\mbs{\mu}^{(n)}_j.
\end{align}
Hence, the family of Gaussian distributions not only makes the algorithm tractable, it simplifies the consensus step by eliminating the normalization involved in equation~\eqref{eq:consensus} by reducing to updates on the mean and covariance matrix. Derivation is provided in the supplementary. 
\end{remarks}

\section{Analytic Results: Rate of Convergence}

\begin{assumption}
\label{assum:network}
The network is a connected aperiodic graph. Specifically, $W$ is an aperiodic and irreducible stochastic matrix.
\end{assumption}

\begin{assumption}
\label{assump:global_learnability}
Let $\Theta$ be a finite set and let $\overline{\Theta}_i := \argmin_{\theta \in \Theta}\expe_{\P_i}\left[D_{\text{KL}}\left(\mc{D}_{Y|X}(\cdot|X_i)||\ell_i(\cdot| \theta, X_i)\right)\right]$ and $\Theta^{\ast} := \cap_{i=1}^N \overline{\Theta}_i$. There exists a parameter $\theta^{\ast} \in \Theta$ that is \textit{globally learnable}, i.e, $\cap_{i=1}^N \overline{\Theta}_i \neq \varnothing $. 
\end{assumption}

\begin{assumption}
\label{assum:bounded_ll}
For all agents $i \in [N]$, assume:(i)   The prior posterior $\est{i}{0} (\theta) > 0$ for all $\theta \in \Theta$. (ii) There exists an $\alpha > 0,\, L > 0$ such that $\alpha < \ell_i(y\mid \theta, x) < L$, for all $y \in \mc{Y}$, $\theta \in \Theta$ and $x \in \mc{X}$.
\end{assumption}

These assumptions are natural. Assumption ~\ref{assum:network} states that one can always restrict attention to the connected components of the social network where the information gathered locally by the agents can disseminated within the component. Assumption ~\ref{assump:global_learnability} ensures the combined observation of the agents across the network is statistically sufficient to learn the global model. Finally, Assumption ~\ref{assum:bounded_ll} prevents the degenerate case where a zero Bayesian prior prohibits learning. 


\begin{theorem}
\label{thm:error_bound}
Let $\Theta$ be a finite set and let $\mc{Q} = \mc{P}(\Theta)$. Under assumptions~\ref{assum:network},~\ref{assump:global_learnability}~and~\ref{assum:bounded_ll}, using the decentralized learning algorithm described in~Sec.~\ref{sec:algorithm} for any given confidence parameter $\delta \in (0,1)$ and any arbitrarily small $\epsilon > 0$, we have 
\begin{align}
\vspace{-0.1cm}
    \max_{i \in [N]}\max_{\theta \not\in \Theta^{\ast}}b_i^{(n)}(\theta) < e^{-n( K(\Theta) - \epsilon)}
\end{align}
when the number of samples satisfies
$
n \geq \frac{8C\log \frac{N |\Theta|}{\delta}}{\epsilon^2 (1-\lambda_{\text{max}}(W))},
$
where we define the rate of convergence of the posterior distribution as follows
\begin{align}
\label{eq:rate_of_conv}
\vspace{-0.1cm}
K(\Theta):= \min_{\theta^{\ast} \in \Theta^{\ast}, \theta \in \Theta \setminus \Theta^{\ast}}\sum_{j = 1}^{N} v_j I_j(\theta^{\ast}, \theta),
\vspace{-0.2cm}
\end{align}
and 
$
I_j(\theta^{\ast}, \theta) 
:=  \expe_{\P^M_j}\left[ D_{\text{KL}}\left( \mc{D}_{Y|X}(\cdot|\mbf{X}_j)|| \ell_j(\cdot \mid \theta, \mbf{X}_j)\right) - D_{\text{KL}}\left( \mc{D}_{Y|X}(\cdot| \mbf{X}_j)|| \ell_j(\cdot\mid \theta^{\ast}, \mbf{X}_j)\right) \right],
$
where eigenvector centrality $\mbf{v} = [v_1, v_2, \ldots, v_N]$ is the unique stationary distribution of $W$ with strictly positive components, furthermore define $\lambda_{\text{max}}(W) := \max_{1\leq i \leq N-1}\lambda_i(W)$,  where $\lambda_i(W)$ denotes $i$-th eigenvalue of $W$ counted with algebraic multiplicity and $\lambda_0(W) = 1$, and $C := \left|\log \frac{L}{\alpha}\right|$. 
\end{theorem}

Proof of the theorem and additional comments on the rate of convergence are provided in the supplementary material.


\begin{remarks}
\label{remm:rate_conv}
The rate of convergence  characterized by~\eqref{eq:rate_of_conv} is a function of the agent’s ability to distinguish between the parameters
given by the KL-divergences and structure of the weighted
network which is captured by the eigenvector centrality $\mbf{v}$ of the agents. Hence, every agent influences the rate in two ways. Firstly, if the agent has higher eigenvector centrality (i.e. the agent is centrality located), it has larger influence over the posterior distributions of other agents as a result has a greater influence over the rate of exponential decay as well. Secondly, if the agent has high KL-divergence (i.e highly informative local observations that can distinguish between parameters), then again it increases the rate. If an influential agent has highly informative observations then it boosts the rate of convergence. We will illustrate this through extensive simulations in~Sec.~\ref{sec:experiment}. 
\end{remarks}


\section{Experiments}
\label{sec:experiment}

\subsection{Decentralized Bayesian Linear Regression}
\label{sec:bayesian_linear_reg}

\begin{figure}
    \centering
    \begin{subfigure}[b]{0.3\textwidth}
        \centering
        \includegraphics[width=1.1\textwidth]{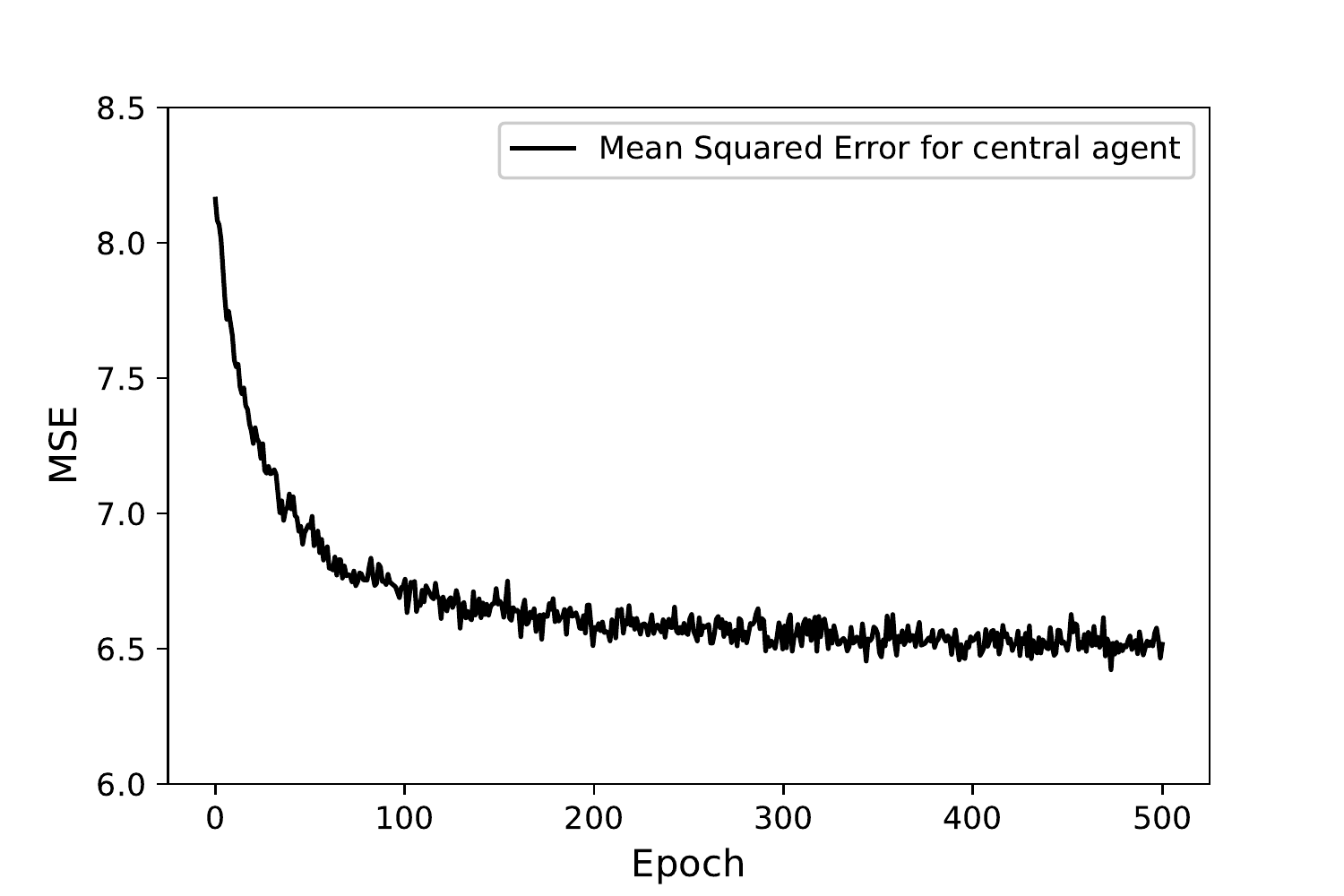}
        \caption{Central agent}
        \label{fig:linear_reg_central}
    \end{subfigure}%
    \hfill
    \begin{subfigure}[b]{0.3\textwidth}
        \centering
        \includegraphics[width=1.1\textwidth]{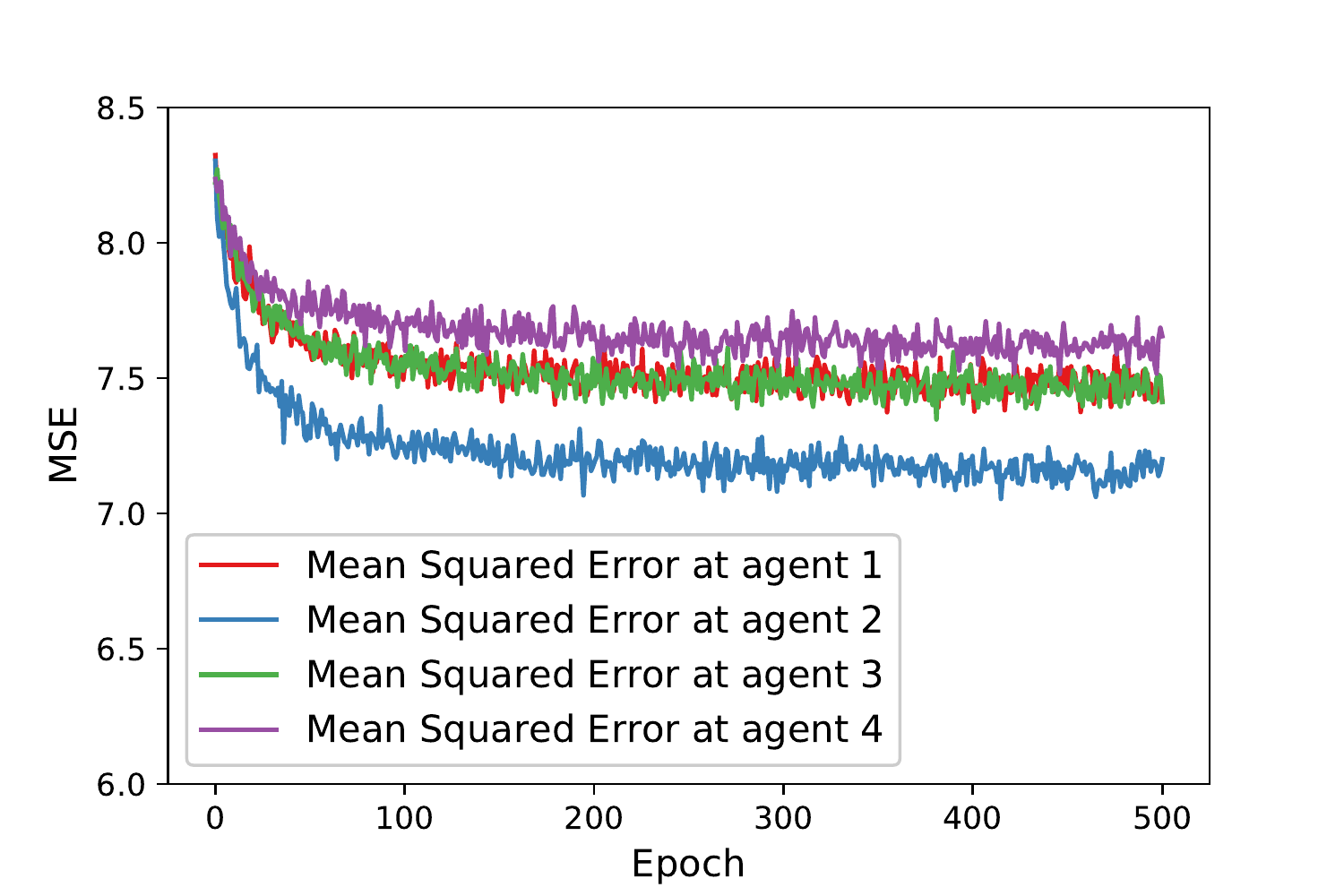}
        \caption{Learning without cooperation}
        \label{fig:linear_reg_nocoop}
    \end{subfigure}
    \hfill
    \begin{subfigure}[b]{0.3\textwidth}
        \centering
        \includegraphics[width=1.1\textwidth]{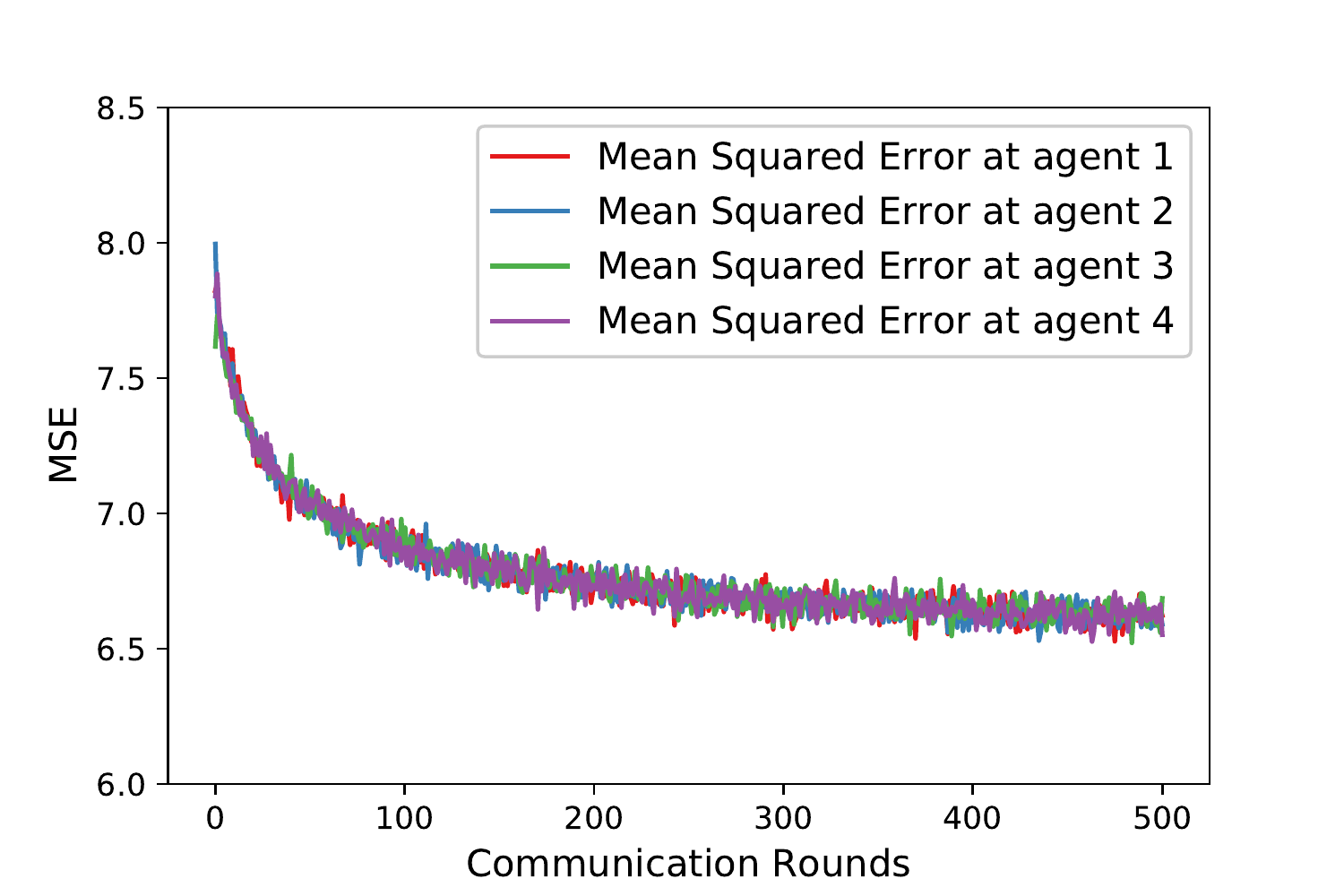}
        \caption{Learning with cooperation}
        \label{fig:linear_reg_coop}
    \end{subfigure}
    \setlength{\belowcaptionskip}{-15pt}
    \caption{Figure compares the Mean Squared Error (MSE) of the predictions over a test dataset under three cases: (i) a benchmark scenario where all training data is shared with a central (cloud) agent, (ii) another benchmark case in which local agents, despite the severe deficiency of their observations, learn without cooperation using local training data only, and (iii) our learning paradigm where agents learn using the proposed decentralized learning rule.}
\label{figure1}
\end{figure}

To illustrate our approach, we construct an example of Bayesian linear regression (Example~\ref{ex:dist_linear_regression}) in the realizable setting over the network with 4 agents. We show that our proposed social learning framework enables a fully decentralized and fast learning of a global model even when the local data is severely deficient. More specifically, we assume that each agent makes observations along only one coordinate of $\mbf{x}$ even though the global test set consists of observations belonging to any $\mbf{x}$ (further details of the experimental setup are provided in the supplementary). Note that this is a case of extreme non-IID data partition across the agents. Fig.~\ref{fig:linear_reg_coop} shows that the MSE of both agents, when trained using the learning rule, matches that of a central agent implying that the agents converge to the true $\mbs{\theta}^{\ast}$ as our theory predicts.




\begin{remarks}
Note that Gaussian likelihood functions considered in Example~\ref{ex:dist_linear_regression} violate the bounded likelihood functions assumption. Furthermore, the parameters belong to a continuous parameter set $\Theta$. This example and those that follow demonstrate that our analytical assumptions on the likelihood functions and the parameter set are sufficient but not necessary for convergence of our decentralized learning rule.
\end{remarks}

\vspace{-0.5cm}
\subsection{Decentralized Image Classification}
\label{sec:bayesian_DNN}


To illustrate the performance of our learning rule on real datasets we consider the problem of decentralized training of Bayesian neural networks for an image classification task on the MNIST digits dataset~\cite{lecun-mnisthandwrittendigit-2010} and the Fashion-MNIST (FMNIST) dataset~\cite{xiao2017_online}. For all our experiments we consider a fully connected NN with the same architecture considered in the context of federated learning in~\cite{mcmahan2017communication}. Additional details regarding the implementation are provided in the supplementary. At each time step $n$, we sample $\mbs{\theta}_{k} \sim \mbf{b}_i^{(n)}$ for $k \in [L]$ and for each test set image $\mbf{x}$, we employ Monte Carlo to obtain the prediction and confidence in the prediction as 
$y = \argmax_{y' \in \mc{Y}} \frac{1}{L} \sum_{k = 1}^L   \text{Softmax}(y',\mbf{f}_{\mbs{\theta}_{k}}(\mbf{x})),$
and $ \P(y) = \frac{1}{L} \sum_{k = 1}^L   \text{Softmax}(y,\mbf{f}_{\mbs{\theta}_{k}}(\mbf{x}))
$ respectively. 
The posterior probability $\P(y)$ in Bayesian Deep Learning literature~\cite{gal2016uncertainty, pmlr-v48-gal16, bayesian_DNN_CV_yarin}, is interpreted as the \textit{confidence} of agent $i$ in predicting $y$ as the true label. In our experiments, we divide the training dataset into subsets with non-overlapping label subsets. Hence, agents must learn $\mbf{b}^{(n)}_i$ such that the resulting predictive distribution can perform well over the global dataset without sharing the local data and hence not having seen input example associated with the labels that are missing locally.
In other words, our agents, at test time, will produce labels of items that they might have never encountered during the training phase.  To make the distinction,
we refer to a label agent $i$ produces as an in-domain (ID) label if training data corresponding to that label is available locally otherwise they are referred to as out-of-domain (OOD) labels. We now describe our empirical studies.




\subsubsection{Design of Social Interaction Matrix $W$} 
\label{sec:eigenvector_centrality}
\begin{figure}
    \centering
    \begin{subfigure}[b]{0.5\textwidth}
        \centering
        \includegraphics[width=0.7\textwidth]{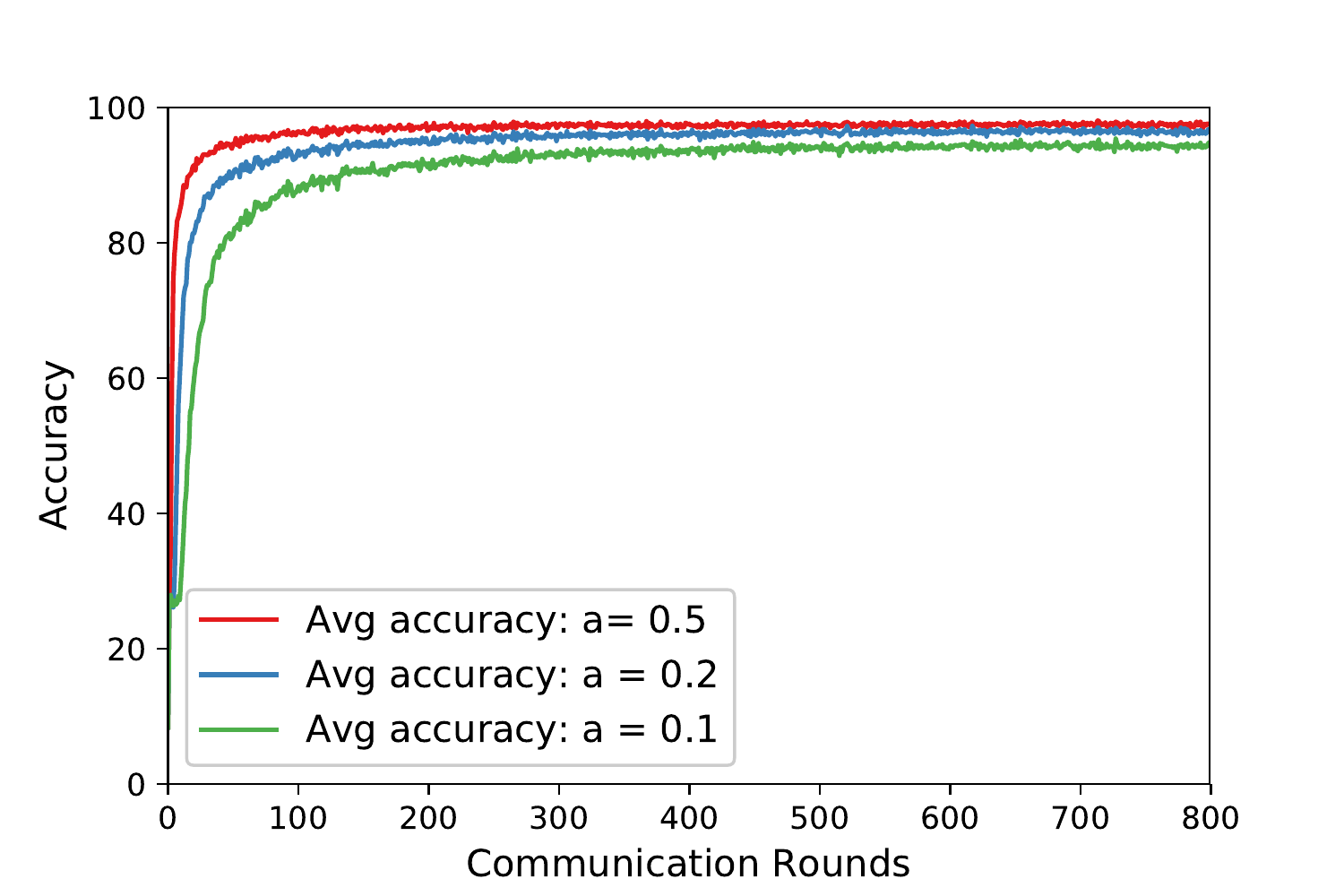}
        \caption{On MNIST dataset}
        \label{fig:9node_star_mnist}
    \end{subfigure}%
    \hfill
    \begin{subfigure}[b]{0.5\textwidth}
        \centering
        \includegraphics[width=0.7\textwidth]{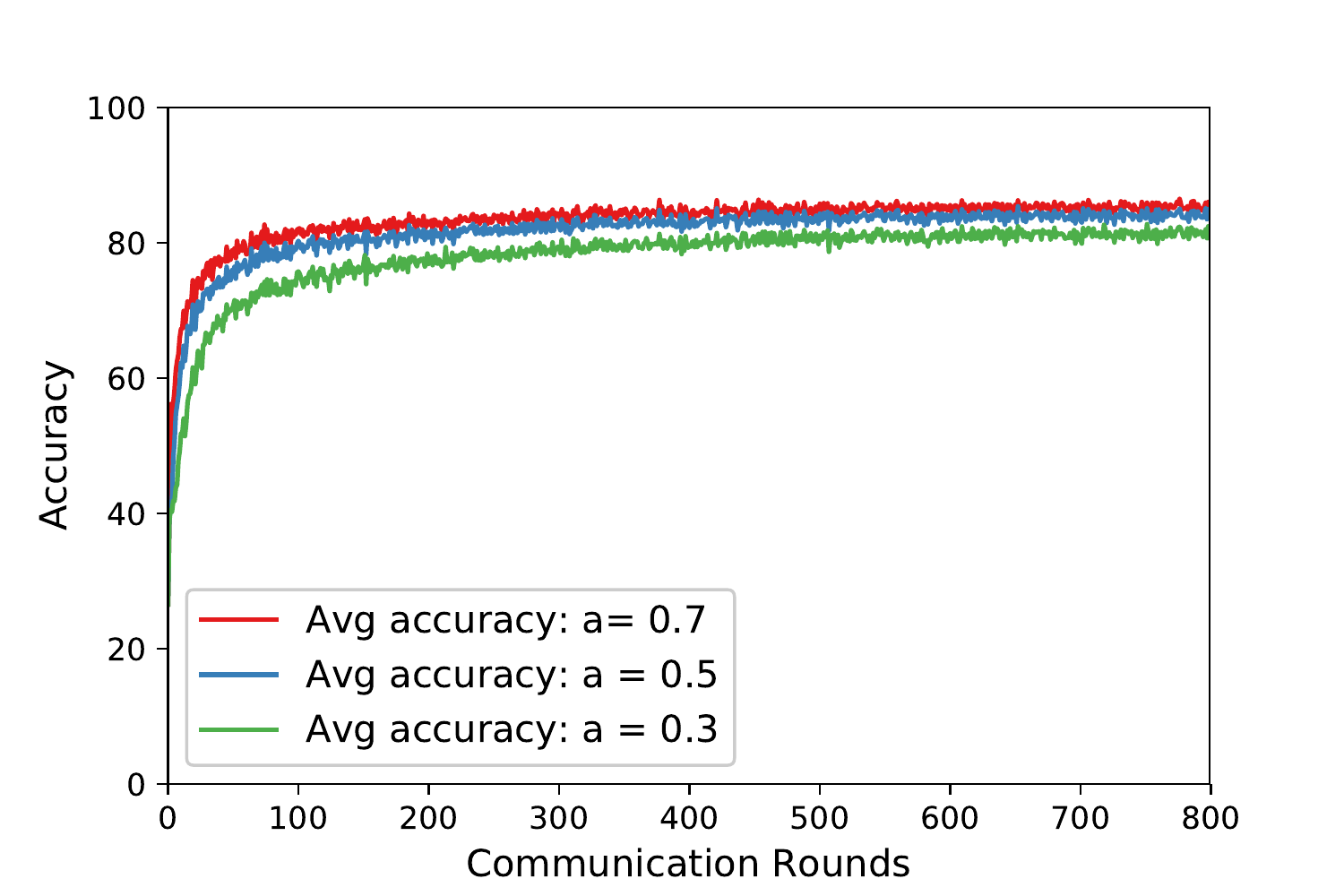}
        \caption{On FMNIST dataset}
        \label{fig:9node_star_fmnist}
    \end{subfigure}
    \setlength{\belowcaptionskip}{-10pt}
    \caption{Figure shows the variation in the average accuracy over a star network topology as the eigenvector centrality of the central agent is varied.}
    \label{fig:9node_star}
\end{figure}

In this section, we investigate how the social interaction matrix $W$ should be designed for a given network structure and a given data partition such that we maximize the rate of convergence in decentralized training. We examine this on a network with a star topology, where a central agent is connected to 8 other edge agents. Let the social interaction weights for the central agent be $\mbf{W}_{1} = [\nicefrac{1}{9}, \ldots, \nicefrac{1}{9}]$. For $a \in (0,1)$, we assume that an edge agent $i$ puts a confidence $\mbf{W}_{i1} = a$ on the central agent, $\mbf{W}_{ii} = 1-a$ on itself and zero on others. Note that as the confidence $a$ which the edge agents put on the central agent increases, the eigenvector centrality of the central agent $v_1$ increases i.e., central agent becomes more influential over the network. For both MNIST and FMNIST, we partition the dataset such that the central agent has more informative local observations. Hence, using equation~\eqref{eq:rate_of_conv} we know that placing more confidence $a$ on the central agent increases the rate of convergence to the true parameter and increases rate of convergence of the test dataset accuracy. This is demonstrated in~Fig.~\ref{fig:9node_star_mnist} and Fig.~\ref{fig:9node_star_fmnist} where both accuracy and the rate of convergence improve as $a$ increases. In other words, rate of convergence and the average accuracy is the highest when the agent with most informative local observations has most influence on the network. Furthermore, on star topology we also demonstrate the scalability of our method through asynchronous implementing over time-varying networks with 25 agents and 100 nodes where we achieve $96.5\%$ and $92.3\%$ accuracy respectively (Sec.~\ref{sec:time_varying} in supplementary). 

We focus on star topology since federated learning methods~\cite{DBLP:journals/corr/KonecnyMRR16, mcmahan2017communication, konevcny2016federated} (only) consider networks with this structure. We compare the performance of our learning rule to the best reported results. On MNIST for $a=0.5$ the average accuracy we obtain is $97.55\%$ which is comparable to the federated learning method~\cite{mcmahan2017communication} FedAvg obtaining $98\%$ for the same architecture and data partition. Similarly, on FMNIST for $a = 0.7$ the average accuracy we obtain is $84.21\%$ slightly inferior to the federated learning method FedAvg~\cite{mcmahan2017communication} which obtains $87.33\%$ accuracy in similar setting. For asynchronous time-varying networks, when we increase the number of agents in the network from 25 to 100, we again see a drop in the accuracy from $96.5\%$ to $92.3\%$ (Sec.~\ref{sec:time_varying} in the supplementary). We believe the lack of periodic global synchronization results in this difference and for detailed discussion refer to Remark~\ref{remm:fed_learning_drawback} in the supplementary. An important area of future work is to overcome this challenge.





\begin{figure}
    \centering
    \begin{subfigure}[b]{0.3\textwidth}
        \centering
        \includegraphics[width=1.1\textwidth]{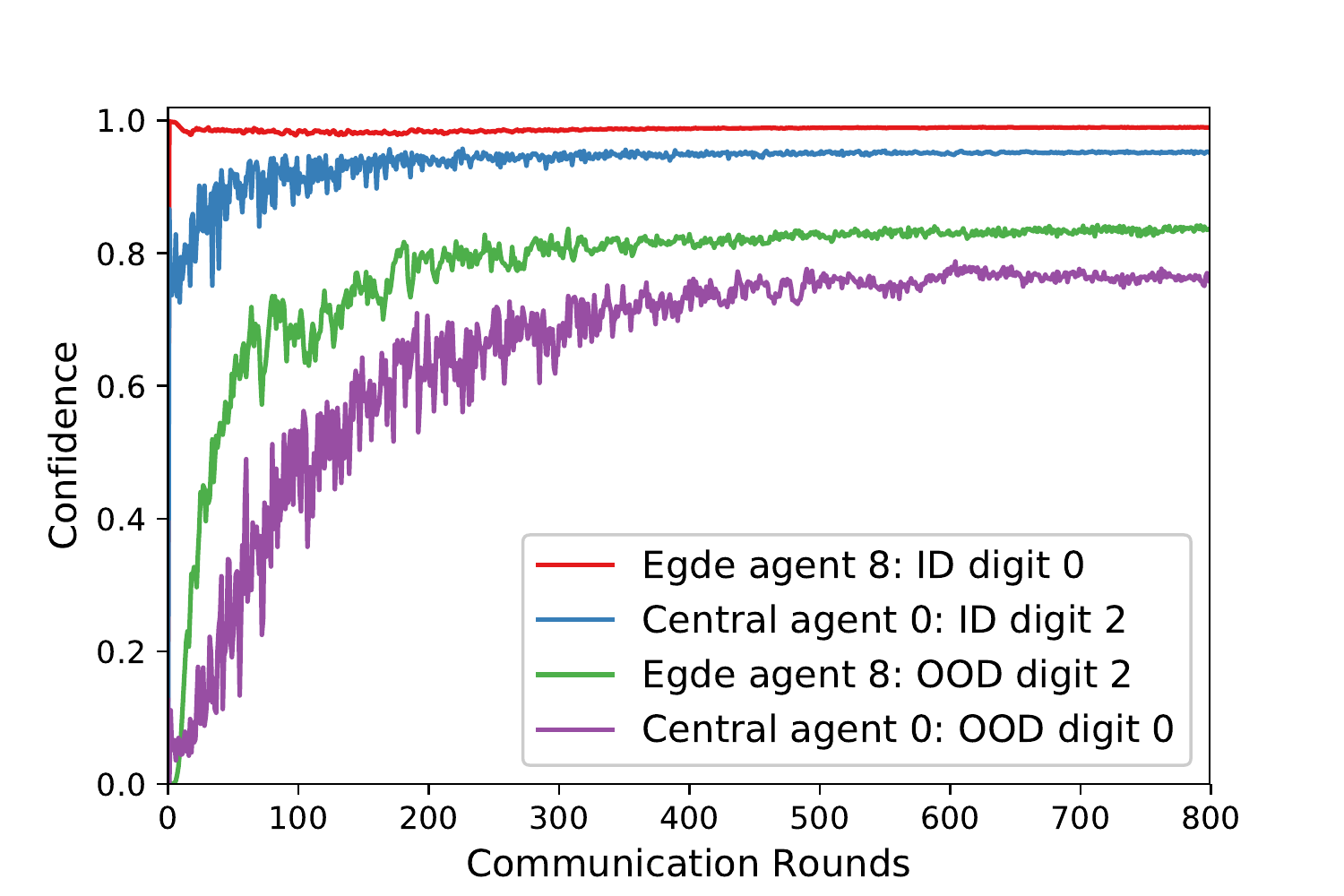}
        \caption{a = 0.1}
        \label{fig:uncert_a_0_1_mnist}    
    \end{subfigure}%
    \hfill
    \begin{subfigure}[b]{0.3\textwidth}
        \centering
        \includegraphics[width=1.1\textwidth]{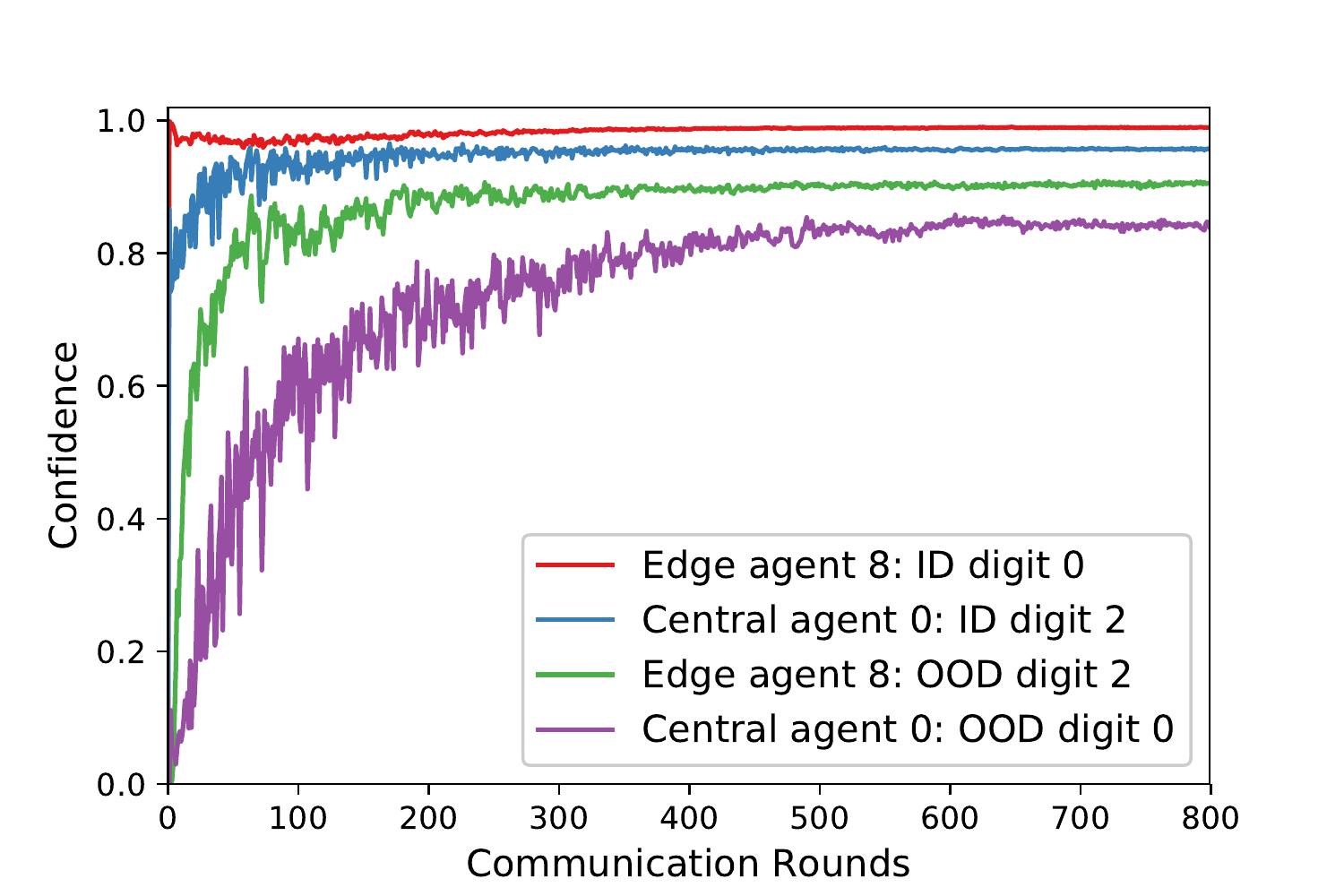}
        \caption{a = 0.2}
        \label{fig:uncert_a_0_2_mnist}
    \end{subfigure}
    \hfill
    \begin{subfigure}[b]{0.3\textwidth}
        \centering
        \includegraphics[width=1.1\textwidth]{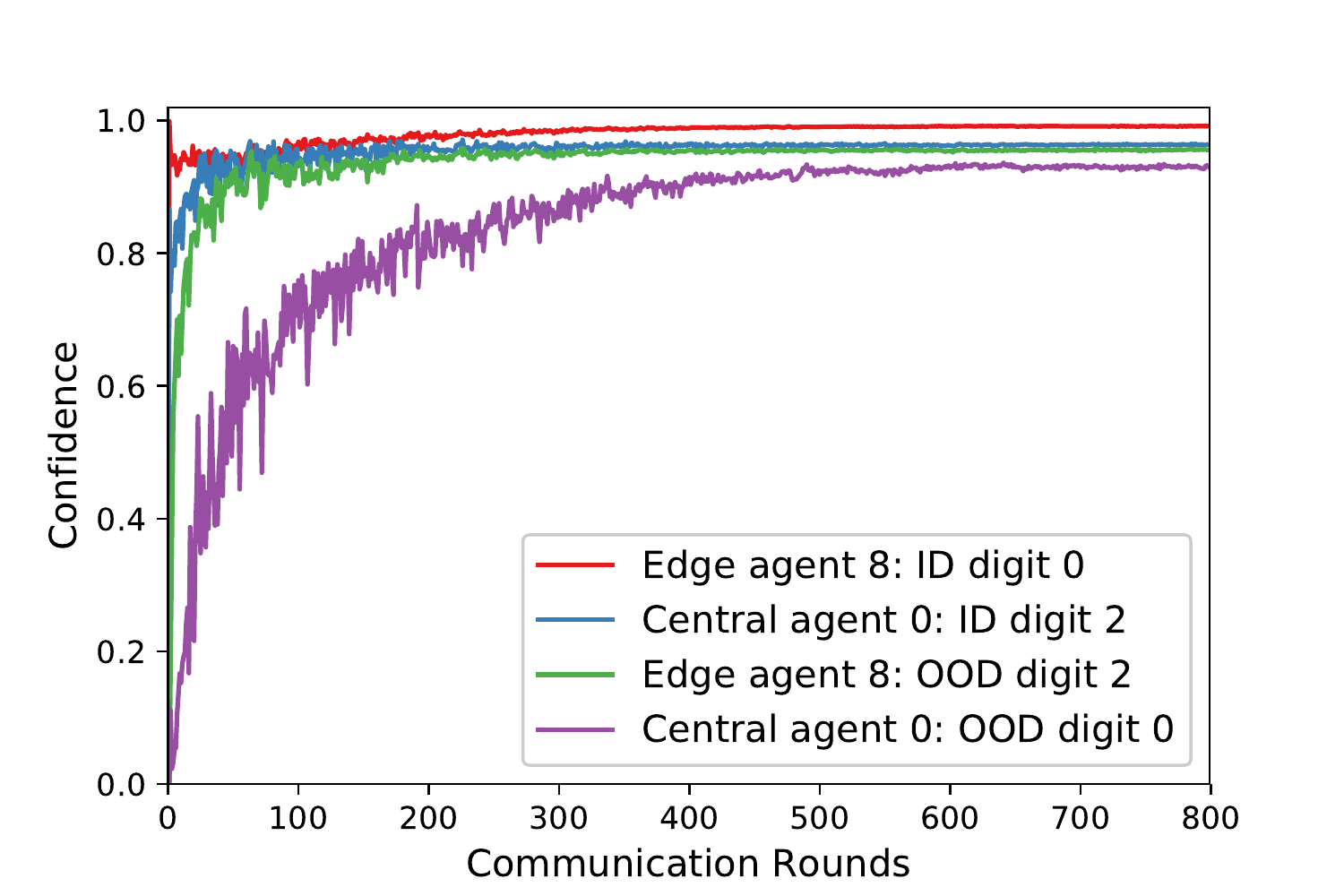}
        \caption{a = 0.5}
        \label{fig:uncert_a_0_5_mnist}
    \end{subfigure}
    \setlength{\belowcaptionskip}{-10pt}
    \caption{Figure shows the increase in the confidence on ID digit and OOD digit at the central agent and an edge agent over communication rounds. Agents are connected in a network with star topology.}
    \label{fig:uncert_mnist}
\end{figure}

\textbf{Effect on confidence over predictions:} In addition to accuracy, Bayesian neural networks provide confidence estimates for each agent's predictions. Hence, we investigate the effect of network structure on confidence. Fig.~\ref{fig:uncert_mnist} shows the confidence on digits 0 and 2 at both central and edge agents as $a$ is varied. In all cases, we observe that both central agent and edge agents learn to predict their ID labels with higher confidence than the OOD labels. Furthermore, Fig.~\ref{fig:uncert_a_0_1_mnist},~Fig.~\ref{fig:uncert_a_0_2_mnist} and Fig.~\ref{fig:uncert_a_0_5_mnist} show that as eigenvector centrality of central agent (most informative agent) increases, the confidence on OOD label at agent edge increase as expected. 



\subsubsection{Effect of Data Partition Over the Network}
\label{sec:local_kldiv}

\begin{figure}
    \centering
    \begin{subfigure}[t]{0.5\textwidth}
        \centering
        \includegraphics[width=0.8\textwidth]{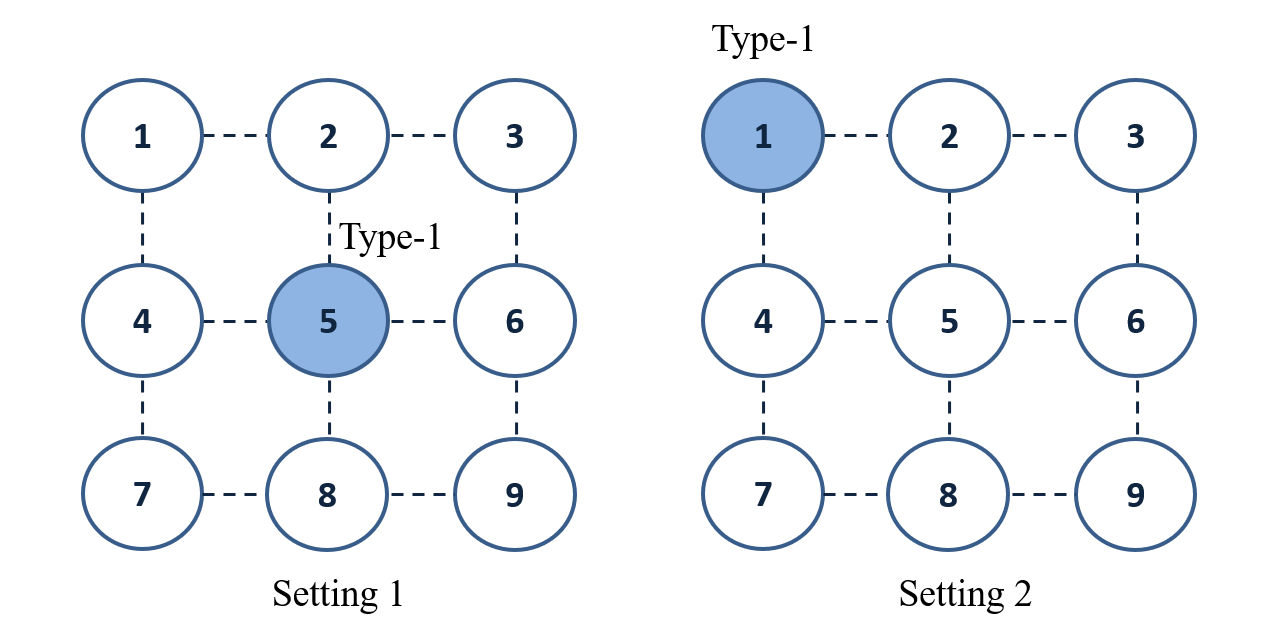}
        \caption{Settings of Grid Topology}
        \label{fig:grid_settings}
    \end{subfigure}%
    \hfill
    \begin{subfigure}[t]{0.5\textwidth}
        \centering
        \includegraphics[width=0.7\textwidth]{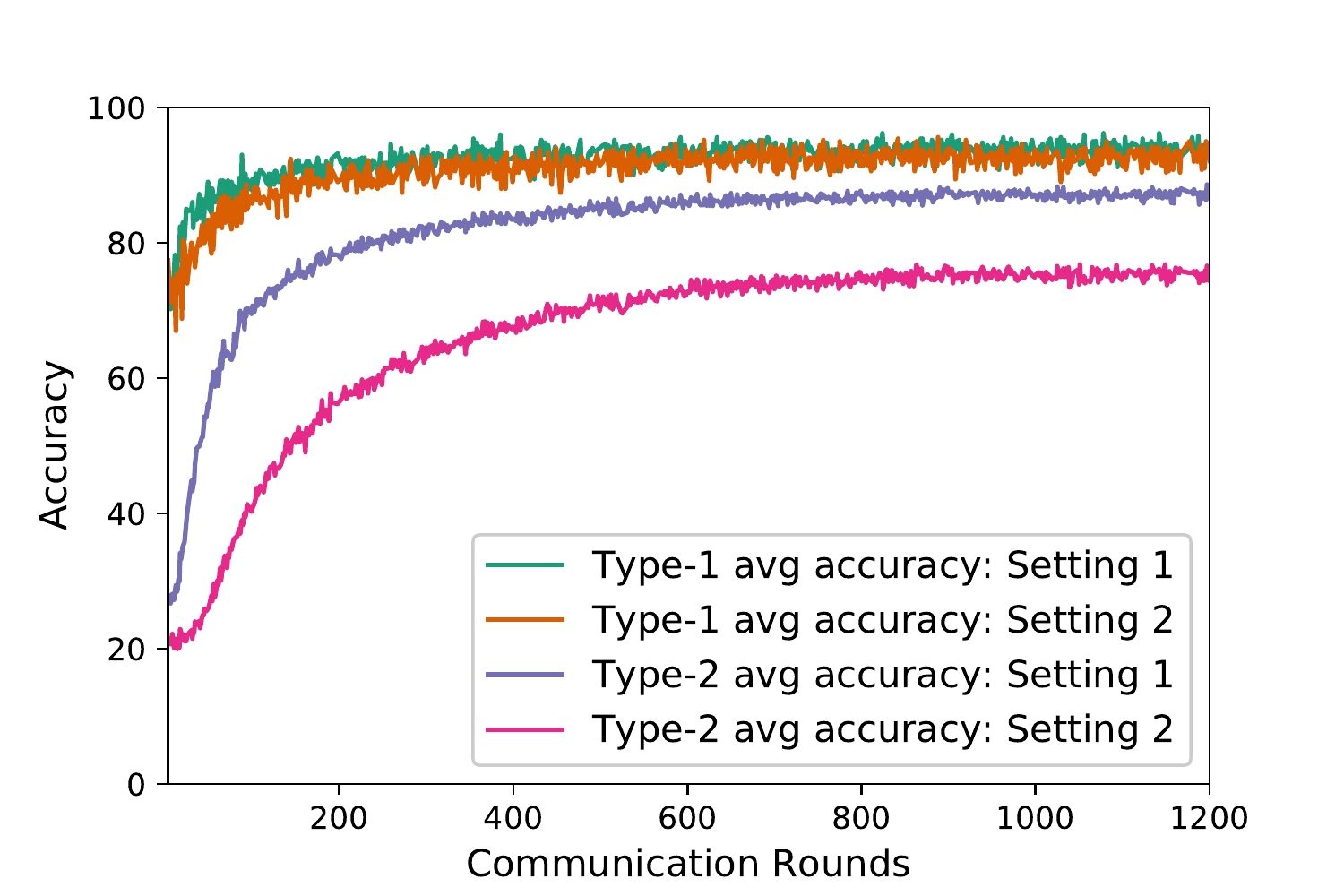}
        \caption{Average accuracy}
        \label{fig:grid_acc}
    \end{subfigure}
    \setlength{\belowcaptionskip}{-10pt}
    \caption{Figure shows average accuracy over 9 agents in a network with grid topology.}
    \label{fig:grid_plots}
\end{figure}

\textbf{Effect of the agent placements:} In this section, we investigate the appropriate placement of a locally informative agent in the network in a manner that maximizes the rate of convergence. We examine this on a $3\times 3$ grid network obtained by connecting every agent to its adjacent agents as shown in Fig.~\ref{fig:grid_settings}. The social interaction weights are defined as $W_{ij} = \nicefrac{1}{|\mc{N}(i)|}$ if $j \in \mc{N}(i)$ and zero otherwise. In this network, the eigenvector centrality of agent $i$ is proportional to its degree $|\mc{N}(i)|$; hence, more number of neighbors implies higher social influence. We divide the data such that the local training set for one of our agents (the Type-1 agent) is statistically informative than the local training set for all other (Type-2) agents. Now, we consider two possible placements of the Type-1 agent in the network~(shown in Fig.~\ref{fig:grid_settings}): (i) \textit{Setting 1:} Type-1 agent is placed at the center (position 5) of the network
and (ii) \textit{Setting 2:} Type-1 agent is placed in a corner location (position 1) in the network. Using equation~\eqref{eq:rate_of_conv} we can predict that setting 1 has a higher rate of convergence to the true parameter and a higher rate of convergence of the test dataset accuracy compared to setting 2 which is demonstrated in~Fig.~\ref{fig:grid_acc}. In other words, rate of convergence is highest when the most influential agent in the network has access to an informative training dataset.

\textbf{Effect of the type of data partition:} Theorem~1 establishes the convergence of our learning rule under Assumption~\ref{assump:global_learnability}. Theoretical implication of this result is that all agents eventually learn the labeling function that best fits the global data if every wrong parameter labeling function can be eliminated by some agent in the network. In the case where the agents use neural networks, local learning can only learn features discriminative to in-domain labels. Our theoretical result suggests that agents are guaranteed to converge to the correct labeling function only when every pair of OOD labels is distinguished by some agents in the network. This also suggests that some non-IID data partitioning of the labels can lead to convergence to an ambiguous set of labeling functions. This has been also shown to lead to poor accuracy empirically in the federated learning literature~\cite{fed_learning_noniid}. Unlike federated learning, our analytic Bayesian framework allows us to theoretically predict the issue. 

\begin{figure}
    \centering
    \begin{subfigure}[b]{0.3\textwidth}
        \centering
        \includegraphics[width=1.1\textwidth]{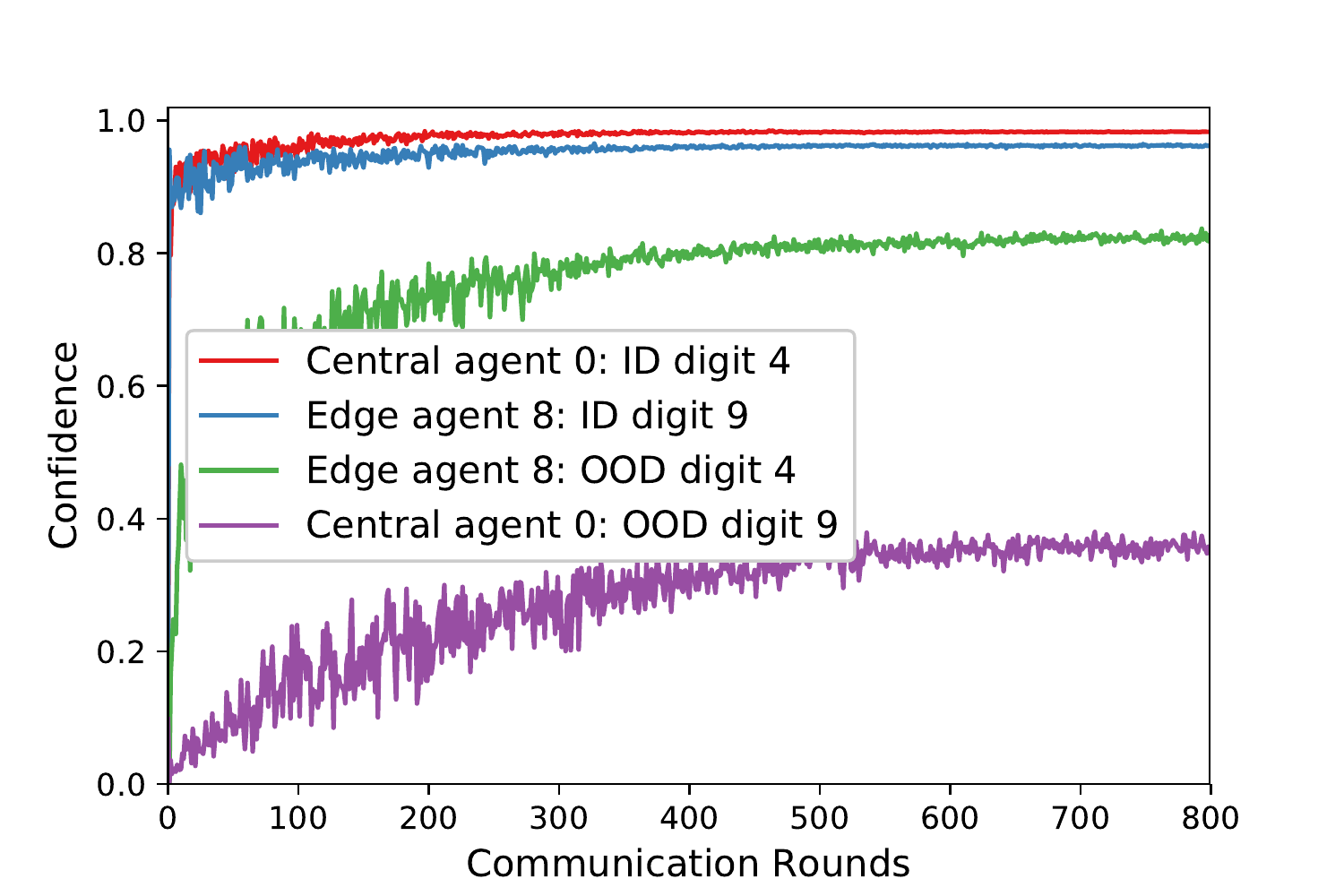}
        \caption{MNIST dataset with ambiguous partitioning}
        \label{fig:mnist_ab_study_case1}
        \end{subfigure}%
    \hfill
    \begin{subfigure}[b]{0.3\textwidth}
        \centering
        \includegraphics[width=1.1\textwidth]{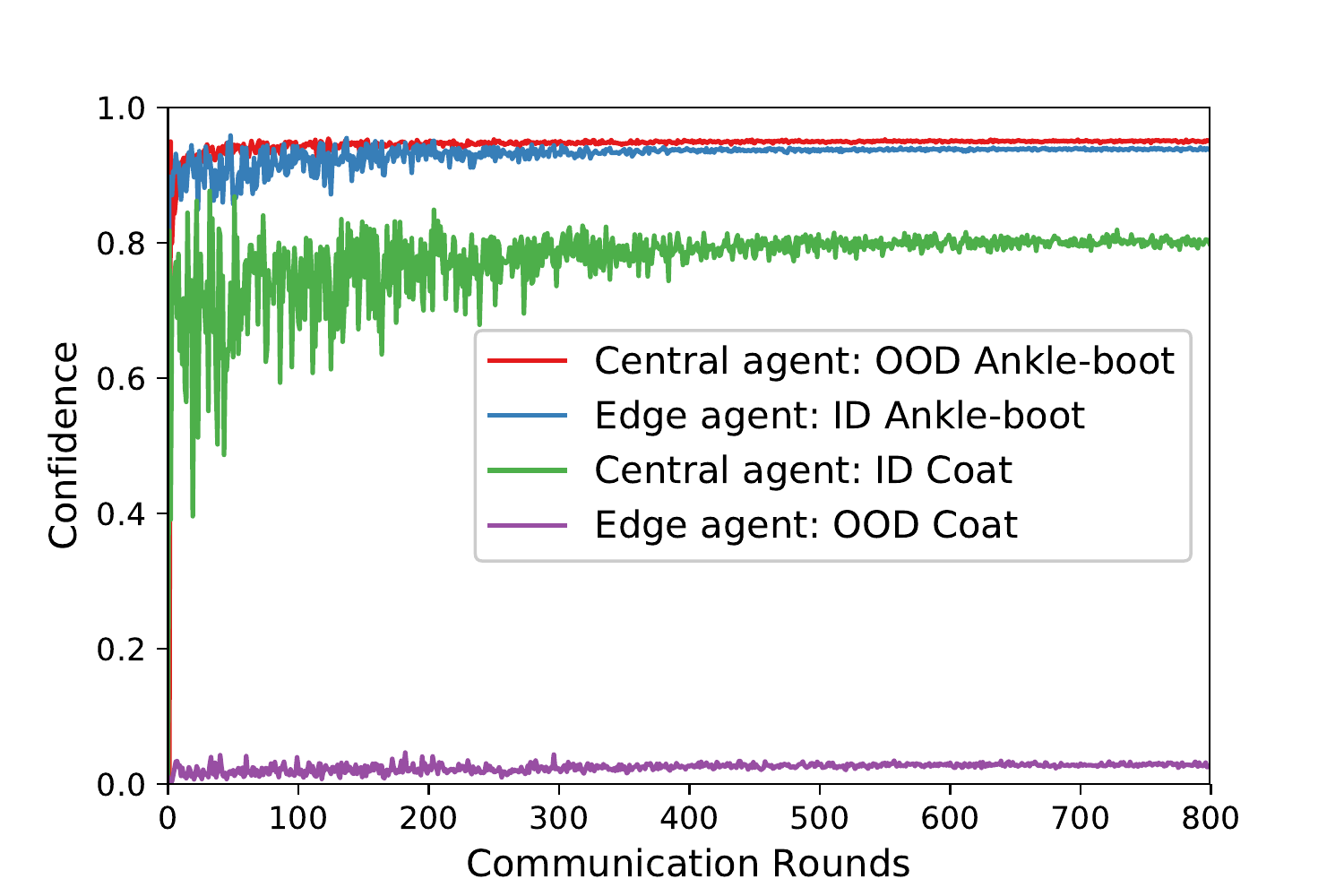}
        \caption{FMNIST dataset with ambiguius partitioning}
        \label{fig:fmnist_ab_study_case1}
    \end{subfigure}
    \hfill
    \begin{subfigure}[b]{0.3\textwidth}
        \centering
         \includegraphics[width=1.1\textwidth]{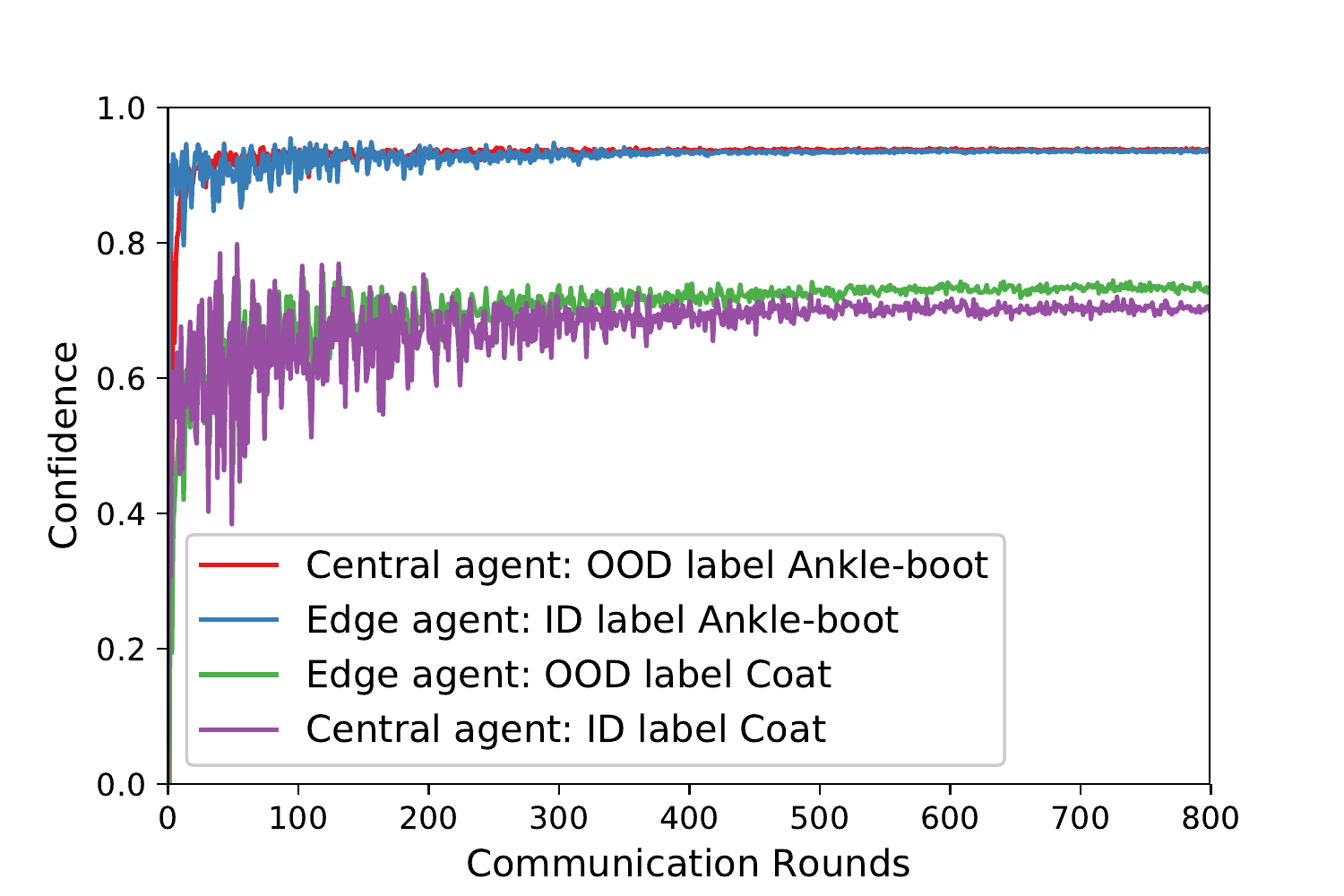}
        \caption{FMNIST dataset with non-ambigius partitioning}
        \label{fig:fmnist_ab_study_case2}
    \end{subfigure}
    \setlength{\belowcaptionskip}{-15pt}
    \caption{Figures shows confidence at central and edge agents in a star network over various partition of the MNIST and FMNIST datasets.}
\end{figure}

In order to understand the practical implications of  Assumption~\ref{assump:global_learnability}, we construct an example where violating assumption leads to poor accuracy. Consider a star network with $a=0.5$ where the central agent has access labels $\{0,\ldots,7\}$ and edge agents have access to labels $\{8,9\}$. Given that $\{4,9\}$ share many common features and since no agent in the network has access to both digits, our analytic results fall short to ensure learning features that can directly distinguish $\{4,9\}$. Indeed, Fig.~\ref{fig:mnist_ab_study_case1} the confidence on OOD digit $9$ at the central agent and on OOD digit $4$ at an edge agent remains low. The effect of data partition described above is more pronounced in the case of FMNIST dataset. Let central agent have access to labels $\{\texttt{t-shirt}$, $\texttt{trouser}$, $\texttt{dress}$, $\texttt{coat}$, $\texttt{shirt}$, $\texttt{bag} \}$ and edge agents has access to labels $\{\texttt{pullover}$,  $\texttt{sandal}$, $\texttt{sneaker}$ $\texttt{ankle-boot}\}$. Agents do not learn to distinguish label $\texttt{pullover}$ at edge agents from the labels at the central agent. Fig.~\ref{fig:fmnist_ab_study_case1} shows that the confidence on OOD label $\texttt{coat}$ at the edge agents is significantly low for this data partition and the average accuracy drops to $69.7\%$. Contrast this with the less ambiguous and less severe data partition of FMNIST data considered for~Fig.~\ref{fig:9node_star_fmnist} where all the labels with shirt-like features, are assigned to a single type, both accuracy and confidence improve as seen in~Fig.~\ref{fig:fmnist_ab_study_case2}.

\section{Conclusion}
In this paper, we considered the problem of decentralized learning over a network of agents with no central server. We considered a peer-to-peer learning algorithm in which agents iterate and aggregate the beliefs of their one-hop neighbors and collaboratively estimate the global optimal parameter. We obtained high probability bounds on convergence and a full characterization of the rate of convergence across the network. We illustrated the effectiveness of algorithm for learning neural networks in computationally tractable manner while achieving high accuracies. Our experimental illustrate the predictive power of analysis of the algorithm. An important area of future work includes extensive empirical studies on various deep neural network architectures.

 \bibliographystyle{unsrt}
\bibliography{ref}

\clearpage
\newpage
\setcounter{page}{1}
\setcounter{section}{1}

\section*{Supplementary Material for Decentralized Bayesian Learning over Graphs}

\subsection{Comments on Rate of Convergence}

\begin{remarks}[Positivity of $K(\Theta)$]
We make a few comments on the quantity $K(\Theta)$. Note that in the realizable setting, for any $\theta^{\ast} \in \Theta^{\ast}$ and $\theta \in \Theta \setminus \Theta^{\ast}$ we get $I_j(\theta^{\ast}, \theta) = \expe_{\P^M_j}\left[ D_{\text{KL}}\left( \ell_j(\cdot|\theta^{\ast}, \mbf{X}_j)|| \ell_j(\cdot \mid \theta, \mbf{X}_j)\right)\right]$ which is non-negative. The KL-divergence between the likelihood functions conditioned on the input captures the extent of distinguishability of parameter $\theta^{\ast}$ from $\theta$. For a wrong parameter $\theta \in \Theta \setminus \Theta^{\ast}$, if $I_j(\theta^{\ast}, \theta)$ is very small then we say that the local observations at agent $j$ are not informative enough to distinguish between $\theta^{\ast}$ and $\theta$. Similarly for the non-realizable setting, for  $\theta^{\ast} \in \Theta^{\ast}$ and $\theta \in \Theta \setminus \Theta^{\ast}$ by definition we have $\expe_{\P^M_i}[D_{\text{KL}}\left( \P_{Y|X}(\cdot| \mbf{X}_j)|| \ell_j(\cdot\mid \theta^{\ast}, \mbf{X}_j)\right)] < \expe_{\P^M_i}[D_{\text{KL}}\left( \P_{Y|X}(\cdot| \mbf{X}_j)|| \ell_j(\cdot\mid \theta, \mbf{X}_j)\right)]$ for all $j$. Hence, $K(\Theta)$ is always positive. In the social learning literature, eigenvector centrality $\mbf{v}$ is a measure of social influence of an agent in the network, since each $v_i$ determines the contribution of agent $i$ in the collective network learning rate $K(\Theta)$. 
\end{remarks}

\subsection{Consensus Step on Gaussian distributions}
Let $(\mbs{\mu}^{(n)}_i, \mbs{\Sigma}_i^{(n)})$ denote the mean and the covariance matrix of $\mbf{b}_i^{(n)}$ at agent at $i$ obtained using equation~\eqref{eq:variational_energy_min}. Using equation~\eqref{eq:consensus}, we have
\begin{align}
&\sum_{j=1}^{N} W_{ij} \ln G(\mbs{\theta}, \mbs{\mu}^{(n)}_j, \mbs{\Sigma}^{(n)}_j)
\\
&= 
-{\frac {1}{2}}\sum_{j=1}^{N} W_{ij} \left(({\mbs {\theta} }-{\boldsymbol {\mu }^{(n)}_j})^{\mathrm {T} }{\boldsymbol {\Sigma }^{(n)}_j}^{-1}({\mbs {\theta} }-{\boldsymbol {\mu }^{(n)}_j})\right) - {\frac {1}{2}}\sum_{i=1}^{N} W_{ij} \ln(2\pi)^k |\mbs{\Sigma}^{(n)}_j|
\\
&= -{\frac {1}{2}} \left( \mbs{\theta}^T \sum_{j=1}^{N}  W_{ij}{\mbs{\Sigma}^{(n)}_j}^{-1} \mbs{\theta} + \sum_{j=1}^{N} {\mbs{\mu}^{(n)}_j}^T W_{ij}{\mbs{\Sigma}^{(n)}_j}^{-1} \mbs{\mu}^{(n)}_j\right)
\\
& \hspace{1cm} + {\frac {1}{2}}\left(\sum_{j=1}^{N} {\mbs{\mu}^{(n)}_j}^T W_{ij}{\mbs{\Sigma}^{(n)}_j}^{-1} \mbs{\theta} + \mbs{\theta}^T \sum_{j=1}^{N} W_{ij}{\mbs{\Sigma}^{(n)}_j}^{-1} \mbs{\mu}^{(n)}_i\right) - {\frac {1}{2}}\sum_{j=1}^{N} W_{ij}\ln(2\pi)^k |\mbs{\Sigma}^{(n)}_j|.
\end{align} 
By completing the squares we obtain $\mbf{q}_i^{(n)}$ is Gaussian distribution and we have
\begin{align}
{\mbs{\widetilde{\Sigma}_i}^{(n)}}^{-1} = \sum_{j=1}^{N}  W_{ij}{\mbs{\Sigma}^{(n)}_j}^{-1},
\end{align}
and
\begin{align}
{\mbs{\widetilde{\Sigma}}^{(n)}}_i^{-1}\mbs{\widetilde{\mu}}^{(n)}_i
=\sum_{j=1}^{N} W_{ij}{\mbs{\Sigma}^{(n)}}_j^{-1}\mbs{\mu}^{(n)}_j
 \implies 
\mbs{\widetilde{\mu}}^{(n)}_i
= {\mbs{\widetilde{\Sigma}}_i^{(n)}}\sum_{j=1}^{N} W_{ij}{\mbs{\Sigma}^{(n)}_j}^{-1}\mbs{\mu}^{(n)}_j.
\end{align}

\subsection{Details on Bayesian Linear Regression Experiment}

Let $\mbs{\theta}^{\ast} = [-0.3, \, 0.5, \, 0.5, \, 0.1, \, 0.2]^T$ and let noise be distributed as $\eta \sim \mc{N}(0, \alpha^2)$ where $\alpha = 0.8$. Agent $i$ makes observations $(\mbf{x}, y)$, where $\mbf{x} = [0\ldots, 0, x_i, 0, \ldots,0]^T$ and $x_i$ is sampled from $\text{Unif}[-1,1]$ for $i = 1$, $\text{Unif}[-1.5,1.5]$ for $i = 2$, $\text{Unif}[-1.25,1.25]$ for $i = 3$, and $\text{Unif}[-0.75,0.75]$ for $i = 4$. We assume each agent starts with a Gaussian prior over $\mbs{\theta}$ with zero mean vector and covariance matrix given by $\text{diag}[0.5, 0.5, 0.5, 0.5]$, where diag$(\mbf{x})$ denotes a diagonal matrix with diagonal elements given by vector $\mbf{x}$. The social interaction weights are given as $\mbf{W}_{1} =[0.5, 0.5, 0, 0]$, $\mbf{W}_{2} =[0.3, 0.1, 0.3, 0.3]$, $\mbf{W}_{3} =[0, 0.5, 0.5, 0]$ and $\mbf{W}_{4} =[0, 0.5, 0, 0.5]$. We assume each agent starts with a Gaussian prior over $\Theta$ and hence the posterior distribution after a Bayesian update remains Gaussian. This implies $\mc{Q}$ remains fixed as the family of Gaussian distributions and the consensus step reduces to equation~\eqref{eq:gaussian_consensus_step}.

\subsection{Details on Bayesian Deep Learning Experiments on Image Classification}

We consider two datasets: (i) the MNIST digits dataset~\cite{lecun-mnisthandwrittendigit-2010} where each image is assigned a label in $\mc{Y} = [0,\ldots,9]$ and (ii) the Fashion-MNIST (FMNIST) dataset~\cite{xiao2017_online} where each image is assigned a label in $\mc{Y} = [$ $\texttt{t-shirt}$, $\texttt{trouser}$, $\texttt{pullover}$, $\texttt{dress}$, $\texttt{coat}$, $\texttt{sandal}$, $\texttt{shirt}$, $\texttt{sneaker}$, $\texttt{bag}$, $\texttt{ankle-boot}]$. Both datasets consist of 60,000 training images and 10,000 testing images of size 28 by 28. For all our experiments we consider a fully connected NN with 2-hidden layers with 200 units each using ReLU activations which is same as the architecture considered in the context of federated learning in~\cite{mcmahan2017communication}.

For all the experiments we choose $\mc{Q}$ to be the family of Gaussian mean-field approximate posterior distributions with pdf given by $G(\mbs{\theta}, \mbs{\mu}, \mbs{\Sigma})$, where $\mbs{\Sigma}$ is a strictly diagonal matrix~\cite{weight_uncert_NN, vcl_iclr}. As discussed in~Remark~\ref{remm:VI} this corresponds to performing variational inference to obtain a Gaussian approximation of the local posterior distribution, i.e., minimizing the variational free energy given in equation~\eqref{eq:variational_energy_min} over $\mc{Q}$. While we compute the KL divergence in~\eqref{eq:variational_energy_min} in a closed form, we employ simple Monte Carlo to compute the gradients using Bayes by Backprop~\cite{weight_uncert_NN, local_repara_trick_kingma}.

\begin{remarks}[Prediction on Test Dataset]
\label{remm:pred_testset}
In the absence of cooperation among the agents, each agent $i$ using the Bayes rule only learns the local posterior distribution $\P(\theta| \mbf{X}_i^{n}, \mbf{Y}_i^{n})$ and makes predictions on the test dataset input $\mbf{x}$ using the predictive distribution $\P(y| \mbf{x}) = \int_{\Theta} \ell_i(y\mid \theta, \mbf{x})\P(\theta| \mbf{X}_i^{n}, \mbf{Y}_i^{n}) d \theta$. However at any time step $n$, using the decentralized learning rule each agent $i$ learns a posterior distribution $\mbf{b}_i^{(n)}$ and makes predictions on the test dataset input $\mbf{x}$ using a  predictive distribution $\int_{\Theta} \ell_i(y \mid \theta, \mbf{x}){b}_i^{(n)}(\theta) d \theta$. Applying~Thm.~\ref{thm:error_bound} we see that as the local posterior $\mbf{b}_i^{(n)}$ converges to $\theta^{\ast}$ for each agent $i$, it can locally predict as if was trained on global dataset.
\end{remarks}

\begin{remarks}
\label{remm:fed_learning_drawback}
Federated learning paradigm, unlike our fully decentralized setup, requires a centralized controller to aggregate the local models from each agent. Furthermore, after each round of communication with the central controller, every agent before training initializes its local model with the global model obtained from the central controller. The periodic shared initialization using a global model across the network, while it is a stringent constraint, is required to prevent the averaging performed at the central controller from producing an arbitrarily bad model~\cite{mcmahan2017communication}. Without modelling the correlation between the weights and bias of the agents across the network, different random initialization at each agent can lead to different local minima and result in diverging local models at the agents~\cite{goodfellow_diff_init}. However, modeling of  correlation between the weights and bias of the agents across the network is computationally prohibitive. We overcome this challenge by using shared initialization when the local models are trained for the first time at each agent, however we do not perform this after each communication round. Our method overcomes the need for shared initialization after each communication round by incorporating the global information (on the weights and bias across the agents) in the local training by using the prior $\mbf{q}_i^{(n)}$, obtained locally via the consensus step~\eqref{eq:consensus} at each agent $i$, in the minimization of variational free energy~\eqref{eq:variational_energy_min}. It would be interesting to investigate other shared initialization suitable for decentralized training which addresses the gap in the performance.  
\end{remarks}

\subsubsection{Design of Social Interaction Matrix $W$}
\label{sec:eig vec exp setting}

For experiments in Sec.~\ref{sec:eigenvector_centrality}, we use a network with a star topology, where there is one central agent and 8 edge agents. We vary confidence $a$ which the edge agents put on the central agent over $[0.1, 0.2, 0.3, 0.5, 0.7]$, the eigenvector centrality of the central agent $v_1$ increases as $[0.1, 0.18, 0.25, 0.36, 0.44]$. We partition the MNIST dataset into two subsets so that the central agent dataset has all images of labels $[2,\ldots,9]$ and edge agents has all images of labels $[0,1]$. To ensure all the edge agents has equal number of images, we shuffle the images with labels $[0,1]$ and partition them into 8 non-overlapping subsets. We call this partition $\texttt{MNIST-Setup1}$.

Similarly, for Fashion-MNIST (FMNIST) dataset, we first partition into two subsets so that central agent has access to labels $[\texttt{t-shirt}$, $\texttt{pullover}$, $\texttt{dress}$, $\texttt{coat}$, $\texttt{shirt}$, $\texttt{bag} ]$ and edge agents have access to labels $[ \texttt{trouser}$,  $\texttt{sandal}$, $\texttt{sneaker}$, $\texttt{ankle-boot}]$. We shuffle the images with labels $[ \texttt{trouser}$,  $\texttt{sandal}$, $\texttt{sneaker}$, $\texttt{ankle-boot}]$ and partition them into 8 non-overlapping subsets. We call this partition $\texttt{FMNIST-Setup1}$. 

We ensure that all agents has same number of local updates $u$ per communication round, which is equal to $(\floor*{n_{edge}/B})E$. For the central agent, this means that for each local epoch, the central agent is trained on a random subset of its local dataset, whereas the edge agents use all the local dataset. For all agents, we use Adam optimizer~\cite{Kingma2015AdamAM} with initial learning rate of 0.001 and learning rate decay of 0.99 per communication round.

\begin{center}
\begin{table}[htp]
\scalebox{0.9}{\begin{tabular}{lclclclclclclclclcl}
\cline{1-9}
Experiment & $E$ & $B$ & $u$ & $\eta$ & $\epsilon$ & $n_{center}$ & $n_{edge}$ &comm rounds\\ \cline{1-9}
$\texttt{MNIST-Setup1}$       &        5      &       50     &          155 & 0.001 & 0.99 & 47335 & 1583 & 800 &\\

$\texttt{MNIST-Setup2}$       &        5      &       50     &          145 & 0.001 & 0.99 & 48200 & 1475 & 800 &\\

$\texttt{MNIST-Setup3}$       &        5      &       50     &          145 & 0.001 & 0.99 & 48209 & 1473 & 800 &\\

$\texttt{FMNIST-Setup1}$      &        5      &       100     &          150 & 0.001 & 0.99 & 36000 & 3000 & 800 &\\

$\texttt{FMNIST-Setup2}$      &        5      &       100     &          150 & 0.001 & 0.99 & 36000 & 3000 & 800 &\\

\end{tabular}}
\caption{Settings for Star Topology Network Experiment: $E$ is number of local epochs, $B$ is the local minibatch size, $u$ is the number of local updates per communication round, $\eta$ is the initial learning rate for all agents, $\epsilon$ is the learning rate decay rate, $n_{center}$ is the dataset size of the central agent, $n_{edge}$ is the dataset size of each of the edge agent.}
\end{table}
\end{center}

\subsubsection{Effect of Data Partition over the Network}
\label{sec:local distinguish}

\textbf{Effect of the agent placements:}
We use a 3 by 3 grid network illustrated by Fig.~\ref{fig:grid_settings} in Sec.~\ref{sec:local_kldiv}. We assign MNIST images with labels $[2,\ldots, 9]$ to an agent of Type-1 and divide images with labels $[0,1]$ among 8 agents of Type-2. In $\texttt{Center}$ setting, we place Type-1 agent at the central location. In $\texttt{Corner}$ setting, we place Type-1 agent in a corner location. Similar to Sec.~\ref{sec:eig vec exp setting}, We ensure that all agents has same number of local updates $u$ per communication round, which is equal to $(\floor*{n_{Type2}/B})E$. Again, we use Adam optimizer for all agents.

\begin{center}
\begin{table}[htp]
\scalebox{0.9}{\begin{tabular}{lclclclclclclclclcl}
\cline{1-9}
Experiment & $E$ & $B$ & $u$ & $\eta$ & $\epsilon$ & $n_{Type1}$ & $n_{Type2}$ &comm rounds\\ \cline{1-9}
$\texttt{Corner}$       &        5      &       50     &          155 & 0.001 & 0.99 & 47335 & 1583 & 1200 &\\

$\texttt{Center}$       &        5      &       50     &          155 & 0.001 & 0.99 & 47335 & 1583 & 1200 &\\

\end{tabular}}
\caption{Settings for Grid Topology Network Experiment: $E$ is number of local epochs, $B$ is the local minibatch size, $u$ is the number of local updates per communication round, $\eta$ is the initial learning rate for all agents, $\epsilon$ is the learning rate decay rate, $n_{Type1}$ is the dataset size of the Type-1 agent, $n_{Type2}$ is the dataset size of each of the Type-2 agent.}
\end{table}
\end{center}

\textbf{Effect of the type data partition:}
In ablation study, we again use a star network and consider two other ways of partitioning the MNIST dataset: (1) the central agent dataset has all images of labels $[0,\ldots.,7]$ and edge agents has all images of labels $[8,9]$, we call this $\texttt{MNIST-Setup2}$, and (2) the edge agents has all images of labels [4,9] and the central agent other labels, we call this $\texttt{MNIST-Setup3}$. For FMNIST dataset, central agent has access to images with labels $[\texttt{t-shirt}$, $\texttt{trouser}$, $\texttt{dress}$, $\texttt{coat}$, $\texttt{shirt}$, $\texttt{bag} ]$ and edge agents have access to images with labels $[\texttt{pullover}$,  $\texttt{sandal}$, $\texttt{sneaker}$ $\texttt{ankle-boot}]$, we call this $\texttt{FMNIST-Setup2}$.

\subsubsection{Asynchronous Decentralized Learning on Time-varying Networks Experiment}

\label{sec:time_varying}

\begin{figure}[!htb]
    \centering
    \begin{subfigure}[b]{0.5\textwidth}
        \centering
        \includegraphics[width=\textwidth]{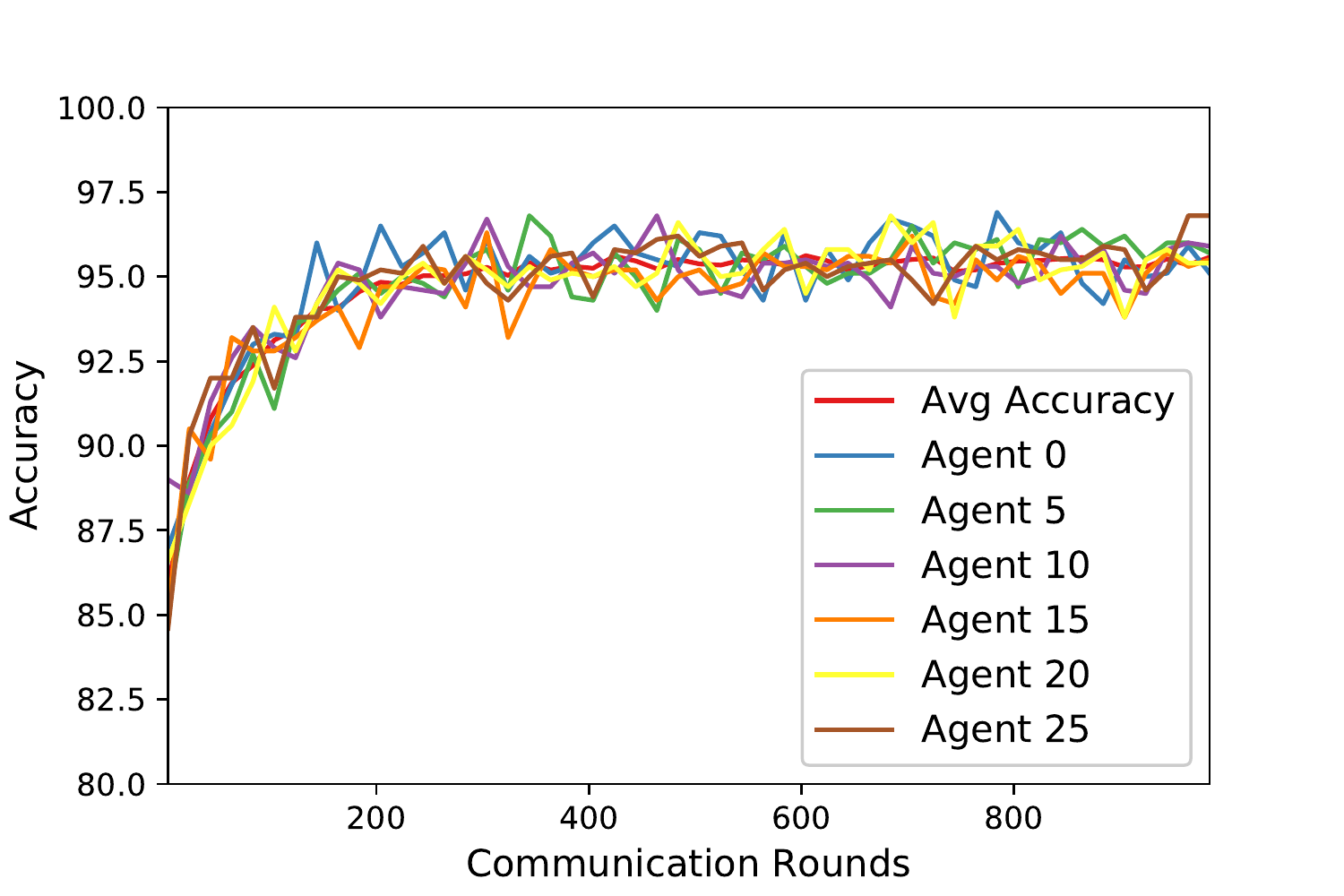}
        \caption{Average accuracy over all 26 nodes.}
    \end{subfigure}%
    \hfill
    \begin{subfigure}[b]{0.5\textwidth}
        \centering
        \includegraphics[width=\textwidth]{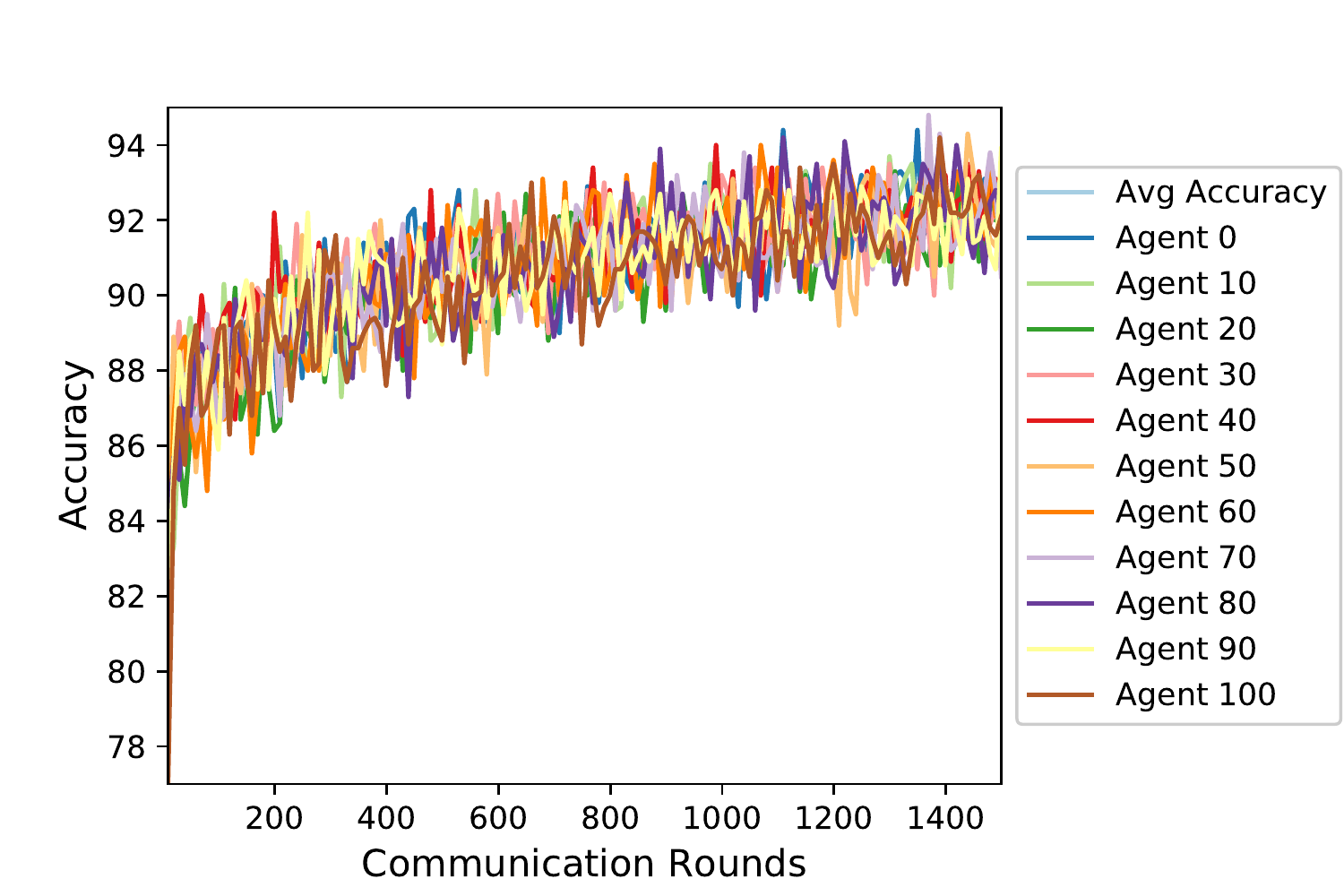}
        \caption{Accuracy over all 100 nodes.}
    \end{subfigure}
    \caption{Figure shows the accuracies of agents in a time-varying network.}
    \label{fig:time_varying_network}
\end{figure}

Now we implement our learning rule on time-varying networks which model practical peer-to-peer networks where synchronous updates are not easy or very costly to implement. We consider a time-varying network of $N+1$ agents numbered as $\{0,1,\ldots, N\}$. At any give time, only $N_0$ agents are connected to agent $0$ in a star topology. For $k \in [\nicefrac{N}{N_0}]$, let $\mc{G}_k$ denote a graph with a star topology where the central agent 0 is connected to edge agents whose indices belong to $\{N_0(k-1)+1, \ldots, N_0 k\}$. This implies at any given time only a small fraction of agents $\nicefrac{N_0}{N}$ are training over their local data. Note that $\cup_{k=1}^{\nicefrac{N}{N_0}} \mc{G}_k$ is strongly connected network over all $N+1$ agents. The social interaction weights for the central agent are $\mbf{W}_{0} = [\nicefrac{1}{N_0+1}, \ldots, \nicefrac{1}{N_0+1}]$. Let $a = 0.5$. An edge agent $i \in \mc{G}_k$ puts a confidence $\mbf{W}_{i0} = a$ on the central agent $0$, $\mbf{W}_{ii} = 1-a$ on itself and zero on others. The MNIST dataset is divided in an i.i.d manner, i.e., data is shuffled and each agent is randomly assigned approximately $(\nicefrac{60,000}{N+1})$ samples. For $N = 25, N_0 = 5$, we obtain an average accuracy of $95.6\%$ over all agents and $95.1\%$ accuracy at the central agent and for $N = 100, N_0 = 10$, we obtain an average accuracy of $92.3\%$ over all agents and $93.1\%$ accuracy at the central agent. This also demonstrates that decentralized learning can be achieved with as few as 600 samples locally.

\begin{center}
\begin{table}[htp]
\scalebox{0.9}{\begin{tabular}{lclclclclclclclcl}
\cline{1-8}
Experiment & $E$ & $B$ & $u$ & $\eta$ & $\epsilon$ & $n$ &comm rounds\\ \cline{1-8}
$\texttt{N = 25}$       &        1      &       50     &          47 & 0.001 & 0.99 & 2307 & 1000 &\\

$\texttt{N = 100}$      &        2      &       10     &          120 & 0.001 & 0.998 & 594 & 1000 &\\

\end{tabular}}
\caption{Settings for Time-varying Network Experiment: $E$ is number of local epochs, $B$ is the local minibatch size, $u$ is the number of local updates per communication round, $\eta$ is the initial learning rate for all agents, $\epsilon$ is the learning rate decay rate, $n$ is the dataset size of any agent. Since all agents have same number of samples, they automatically have equal number of local updates per communication round. Adam optimizer is used for all agents.}
\end{table}
\end{center}

\subsection{Additional Figures}

\begin{figure}[!htb]
    \centering
    \begin{subfigure}[b]{0.5\textwidth}
        \centering
        \includegraphics[width=\textwidth]{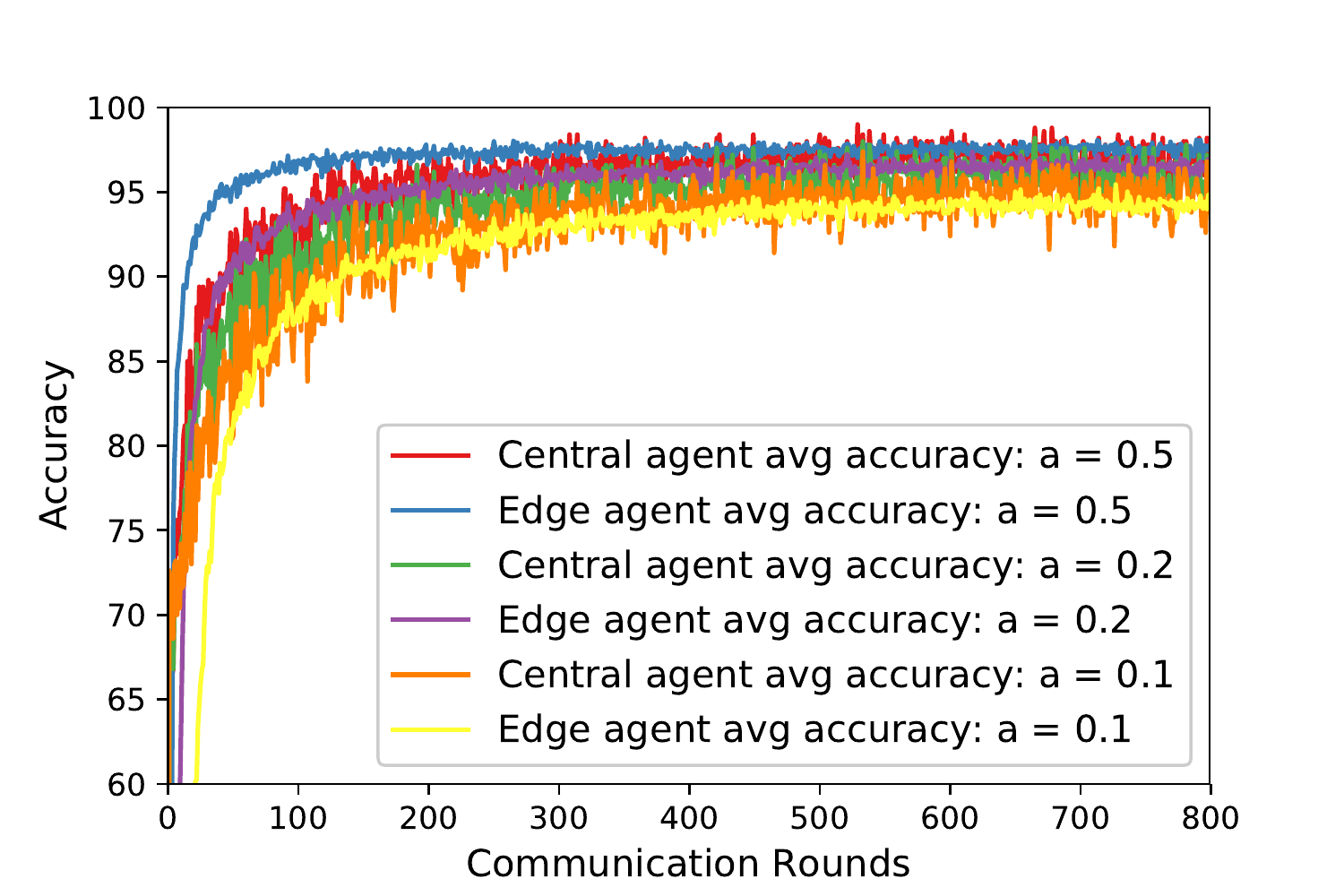}
        \caption{On MNIST dataset}
    \end{subfigure}%
    \hfill
    \begin{subfigure}[b]{0.5\textwidth}
        \centering
        \includegraphics[width=\textwidth]{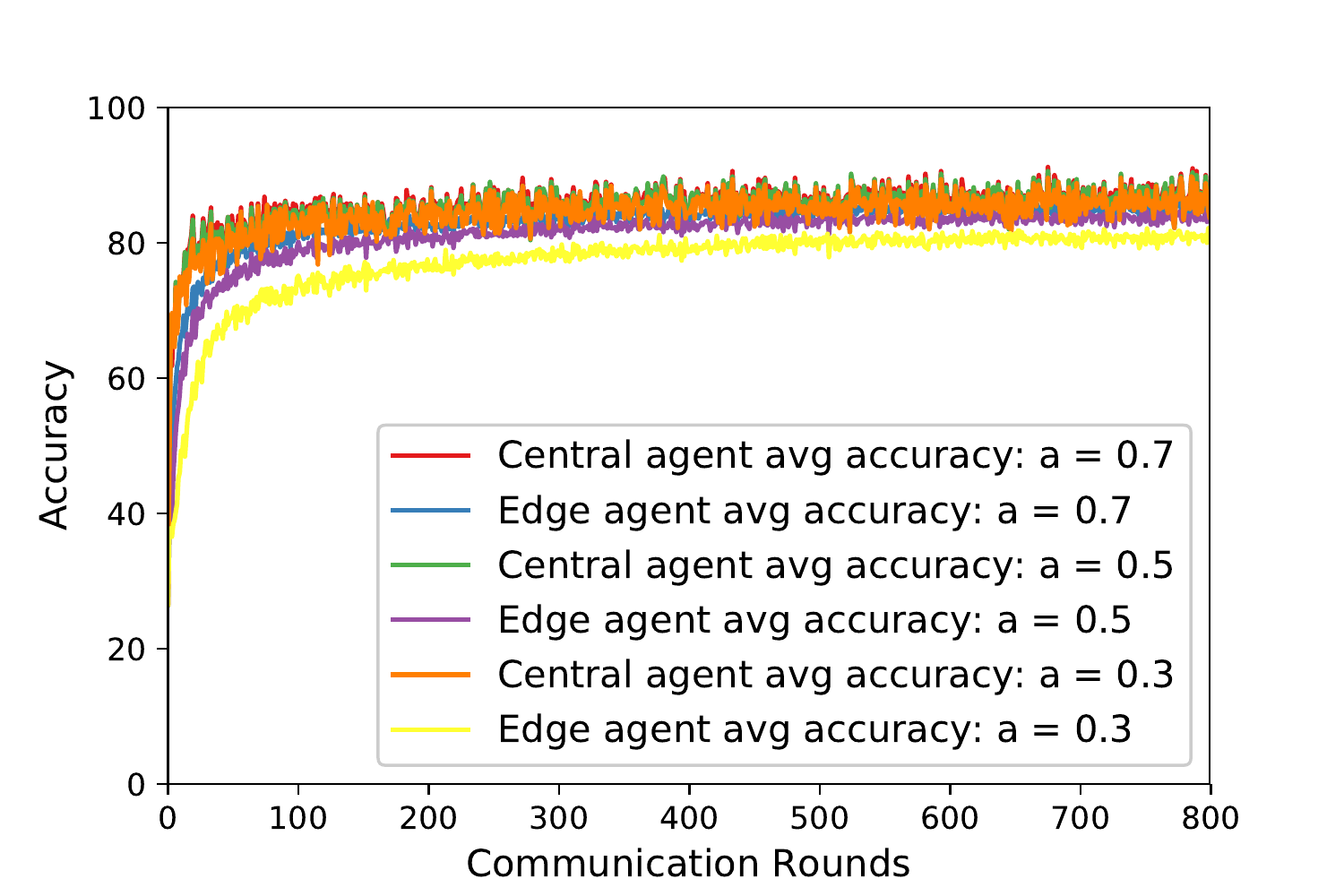}
        \caption{On Fashion-MNIST dataset}
    \end{subfigure}
    \caption{Figures shows average accuracy of 9 agents connected in a network with star topology.}
\end{figure}

\begin{figure}[!htb]
    \centering
    \begin{subfigure}[b]{0.3\textwidth}
        \centering
        \includegraphics[width=1.1\textwidth]{uncert_est_vs_comm_9node_star_a_07_FMNIST}
        \caption{a = 0.7}
    \end{subfigure}%
    \hfill
    \begin{subfigure}[b]{0.3\textwidth}
        \centering
        \includegraphics[width=1.1\textwidth]{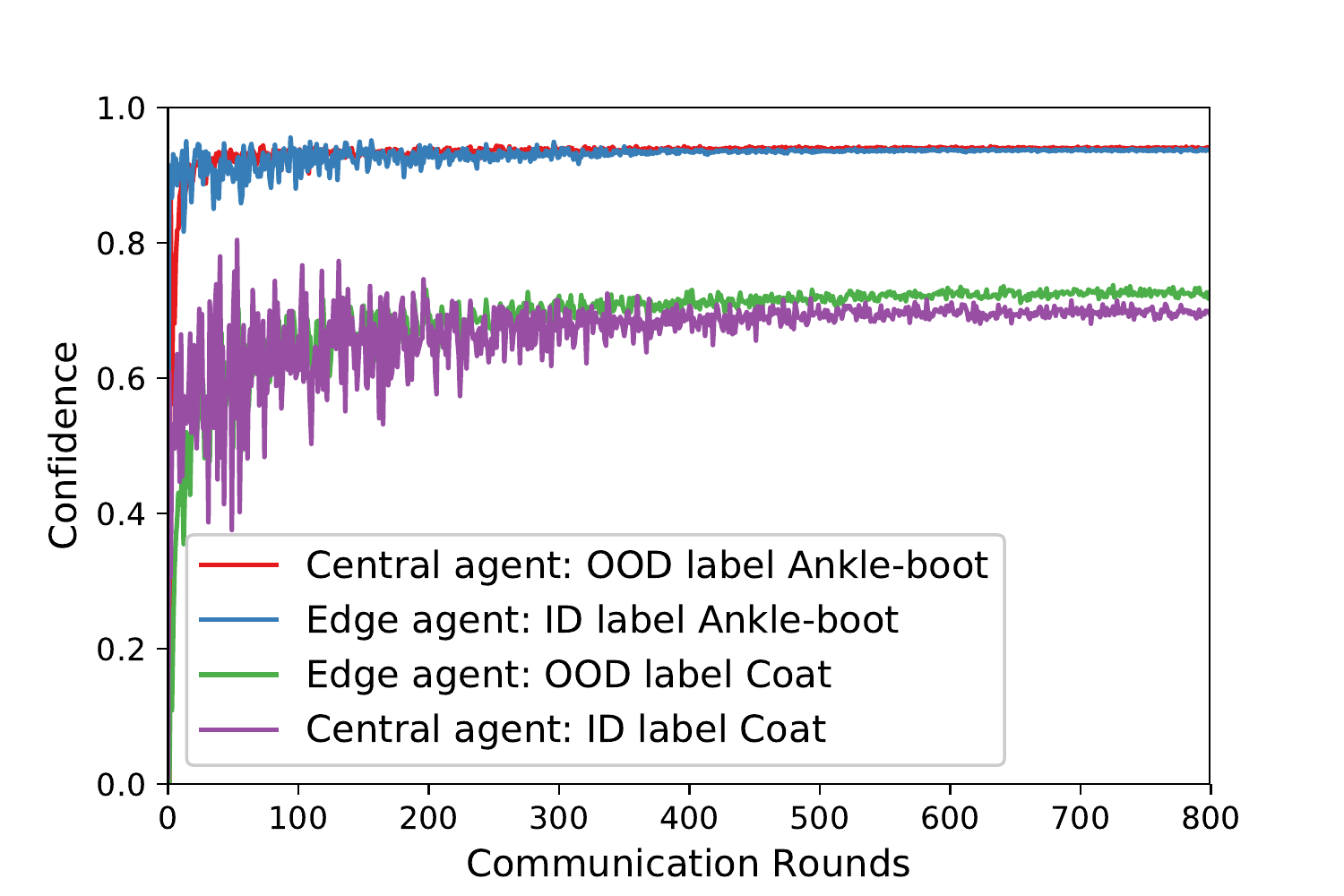}
        \caption{a = 0.5}
    \end{subfigure}
    \hfill
    \begin{subfigure}[b]{0.3\textwidth}
        \centering
        \includegraphics[width=1.1\textwidth]{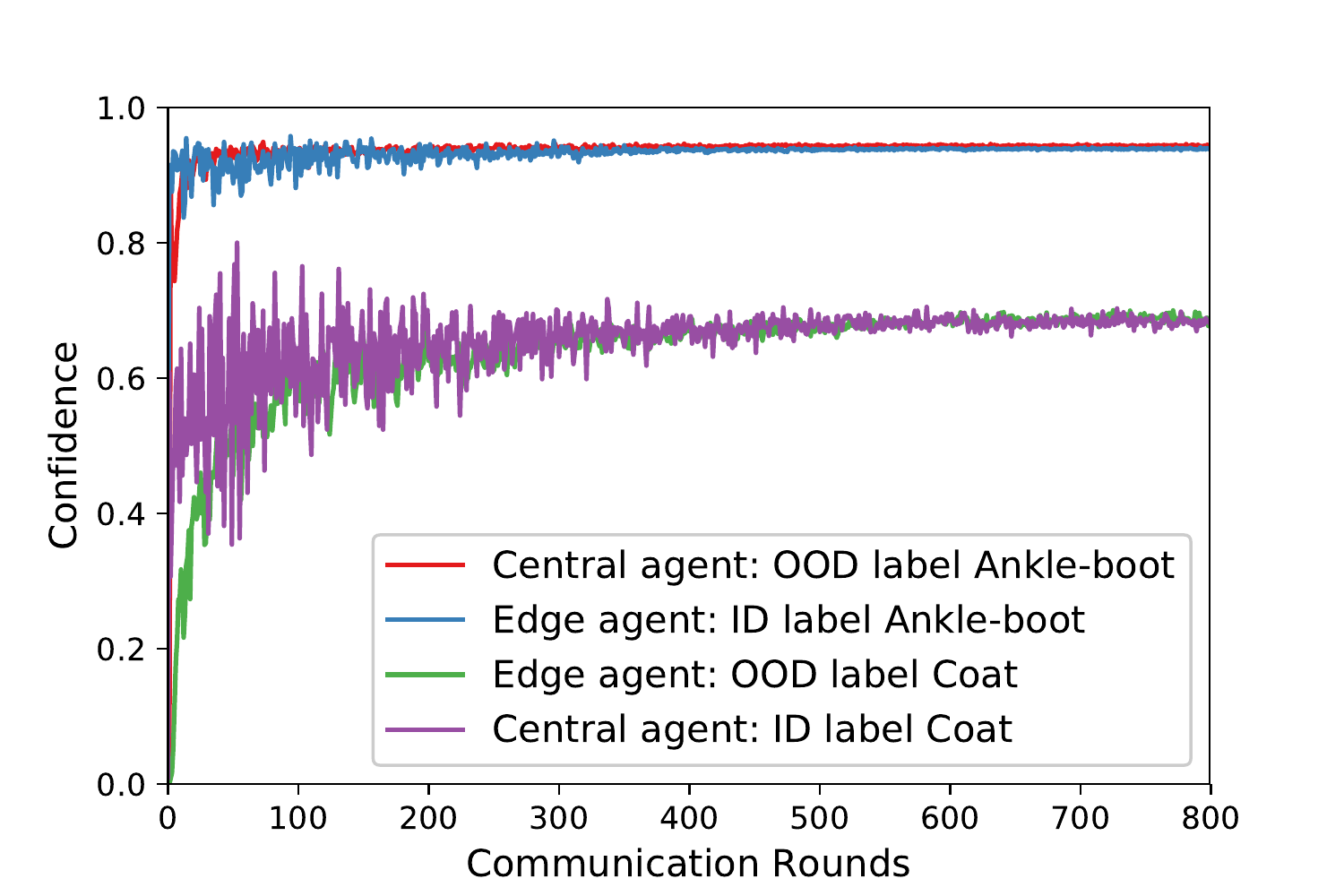}
        \caption{a = 0.3}
    \end{subfigure}
    \caption{Figures shows the increase in the confidence on an ID label and OOD label at the central and edge agents over communication rounds for FMNIST dataset. Agents are connected in a network with star topology and the value of $a$ varies over $[0.7, 0.5, 0.3]$.}
\end{figure}

\begin{figure}[!htb]
    \centering
    \begin{subfigure}[t]{0.5\textwidth}
        \centering
        \includegraphics[width=0.8\textwidth]{avg_acc_T1_T2_vs_comm_9node_grid_MNIST}
        \caption{Average accuracy}
        \label{fig:grid_acc}
    \end{subfigure}%
    \hfill
    \begin{subfigure}[t]{0.5\textwidth}
        \centering
        \includegraphics[width=0.8\textwidth]{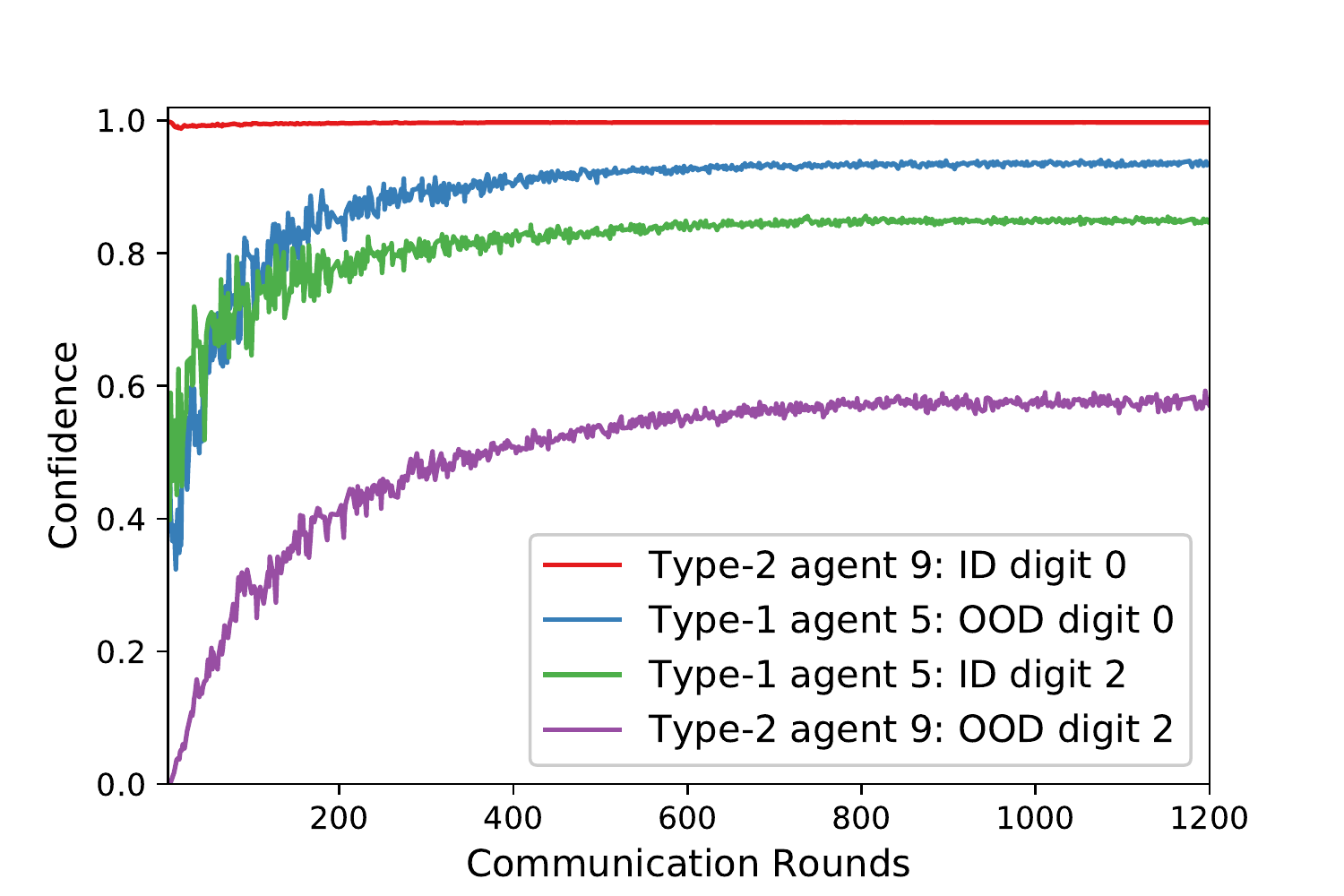}
        \caption{Confidence values in setting 1}
        \label{fig:grid_central_uncert}
    \end{subfigure}
    \caption{Figure shows average accuracy over 9 agents in a network with grid topology.}
    \label{fig:grid_plots}
\end{figure}

\clearpage
\subsection{Proof of Theorem~1}
\label{app:proof}
The proof of Theorem~1 is based the proof provided in~\cite{7172262, 7349151, 8359193}. For the ease of exposition, let $\est{i}{0}(\theta) =\frac{1}{|\Theta|}$ for all $\theta \in \Theta$. Fix a $\theta^{\ast} \in \Theta^{\ast}$. We begin with the following recursion for each node $i \in [N]$ and for any $\theta \not\in \Theta^{\ast}$,
\begin{align}
 & \frac{1}{n}\log \frac{\est{i}{n}(\theta^{\ast})}{\est{i}{n}(\theta)} 
= \frac{1}{n}\sum_{j = 1}^{N}\sum_{k = 1}^{n} W^{k}_{ij}z_j^{(n-k+1)}(\theta^{\ast}, \theta),
 \end{align}
 where
 \begin{align}
z_j^{(k)}(\theta^{\ast}, \theta) = \log \frac{\dist{j}{\samp{j}{k}}{\theta^{\ast}, X_i^{(k)}} }{\dist{j}{\samp{j}{k}}{\theta, X_i^{(k)}}}.
 \end{align}
From the above recursion we have
\begin{align}
\frac{1}{n}\log \frac{\est{i}{n}(\theta^{\ast})}{\est{i}{n}(\theta)}
&\geq 
\frac{1}{n}\sum_{j = 1}^{N} v_j\left(\sum_{k = 1}^{n}  z_j^{(k)}(\theta^{\ast}, \theta)\right)
-\frac{1}{n}\sum_{j = 1}^{N} \sum_{k = 1}^{n} \left| W^{k}_{ij} - v_j \right| \left| z_j^{(k)}(\theta^{\ast}, \theta) \right|
\\
& \overset{(a)}\geq
\frac{1}{n}\sum_{j = 1}^{N} v_j \left(\sum_{k = 1}^{n} z_j^{(k)}(\theta^{\ast}, \theta)\right)-\frac{4C\log N}{n(1-\lambda_{\text{max}}(W))},
\end{align}
where $(a)$ follows from Lemma~\ref{lemma:conv_W} and the boundedness assumption of log-likelihood ratios. Now fix $n \geq \frac{8C\log N}{\epsilon (1-\lambda_{\text{max}}(W))}$, since $\est{i}{n}(\theta^{\ast}) \leq 1$ we have
\begin{align*}
     -\frac{1}{n}\log \est{i}{n}(\theta) 
     \geq
     -\frac{\epsilon}{2}
     +
     \frac{1}{n}\sum_{j = 1}^{N} v_j \left(\sum_{k = 1}^{n} z_j^{(k)}(\theta^{\ast}, \theta)\right).
\end{align*}
Furthermore, we have
\begin{align*}
    &\P\left( -\frac{1}{n}\log \est{i}{n}(\theta)  \leq \sum_{j = 1}^{N} v_j I_j(\theta^{\ast}, \theta) - \epsilon \right)
    \leq 
    \P\left( 
     \frac{1}{n}\sum_{j = 1}^{N} v_j \sum_{k = 1}^{n} z_j^{(k)}(\theta^{\ast}, \theta)  \leq  \sum_{j = 1}^{N} v_j I_j(\theta^{\ast}, \theta) - \frac{\epsilon}{2} \right).
\end{align*}

Now for any $j \in [N]$ note that\vspace{-0.2cm}
\begin{align*}
   \sum_{j = 1}^{N} v_j\sum_{k = 1}^{n}   z_j^{(k)}(\theta^{\ast}, \theta) 
   - n \sum_{j = 1}^{N} v_j I_j(\theta^{\ast}, \theta)
   &= \sum_{k = 1}^{n}   \left( \sum_{j = 1}^{N} v_jz_j^{(k)}(\theta^{\ast}, \theta) - \sum_{j = 1}^{N} v_j\expe[z_j^{(k)}(\theta^{\ast}, \theta)]\right).
\end{align*}
For any $\theta \not\in \Theta^{\ast}$, applying McDiarmid's inequality for all $\epsilon > 0$  and for all $n \geq 1$ we have
\begin{align*}
    &\P\left( \sum_{k = 1}^{n}   \left( \sum_{j = 1}^{N} v_j z_j^{(k)}(\theta^{\ast}, \theta) - \sum_{j = 1}^{N} v_j\expe[z_j^{(k)}(\theta^{\ast}, \theta)]\right)  \leq -\frac{\epsilon n}{2}  \right)
    \leq 
    e^{-\frac{\epsilon^2 n}{2C}}.
\end{align*}

Hence, for all $\theta \not\in \Theta^{\ast}$, for $n \geq \frac{8C\log N}{\epsilon (1-\lambda_{\text{max}}(W))}$ we have
\begin{align}
    \P\left(\frac{-1}{n}\log \est{i}{n}(\theta)  \leq \sum_{j = 1}^{N} v_jI_j(\theta^{\ast}, \theta) - \epsilon\right)  \leq 
    e^{-\frac{\epsilon^2 n}{4C}},
\end{align}
which implies
\begin{align}
    \P\left(\est{i}{n}(\theta) \geq e^{-n( \sum_{j = 1}^{N} v_jI_j(\theta^{\ast}, \theta) - \epsilon)} \right)  \leq 
    e^{-\frac{\epsilon^2 n}{4C}}.
\end{align}
Using this we obtain a bound on the worst case error over all $\theta$ and across the entire network as follows
\begin{align}
    \P\left(\max_{i \in [N]}\max_{\theta \not\in \Theta^{\ast}}\est{i}{n}(\theta) \geq e^{-n( K(\Theta) - \epsilon)} \right)  \leq 
    N |\Theta|e^{-\frac{\epsilon^2 n}{4C}},
\end{align}
where $K(\Theta):= \min_{\theta \in \Theta^{\ast}, \psi \in \Theta\setminus \Theta^{\ast}}\sum_{j = 1}^{N} v_j I_j(\theta, \psi)$. From Assumption~\ref{assump:global_learnability} and Lemma~\ref{lemma:conv_W} we have that $K(\Theta) >0$. Then, with probability $1-\delta$ we have 
\begin{align}
    \max_{i \in [N]}\max_{\theta \not\in \Theta^{\ast}}\est{i}{n}(\theta) < e^{-n( K(\Theta) - \epsilon)},
\end{align}
when the number of samples satisfies
\begin{align}
  n \geq \frac{8C\log \frac{N |\Theta|}{\delta}}{\epsilon^2 (1-\lambda_{\text{max}}(W))}.
\end{align}

\begin{lemma}[~\cite{7349151}]
\label{lemma:conv_W}
For an irreducible and aperiodic stochastic matrix $W$, the stationary distribution $\mbf{v} = [v_1, v_2, \ldots, v_N]$ is unique and has strictly positive components and satisfies $v_i = \sum_{j = 1}^{n}v_j W_{ji}$. Furthermore, for any $i \in [N]$ the weight matrix satisfies\vspace{-0.2cm}
\begin{align*}
    \sum_{k = 1}^n \sum_{j=1}^N \left| W^{k}_{ij} - v_j \right| \leq \frac{4\log N}{1-\lambda_{\text{max}}(W)},
\end{align*}
where $\lambda_{\text{max}}(W) = \max_{i \in [N-1]}\lambda_i(W)$, and $\lambda_i(W)$ denotes eigenvalue of $W$ counted with algebraic multiplicity and $\lambda_0(W) = 1$.
\end{lemma}

\end{document}